\documentclass{article}

\usepackage[main, final]{neurips_2026}

\usepackage[utf8]{inputenc} 
\usepackage[T1]{fontenc}    
\usepackage{hyperref}       
\usepackage{url}            
\usepackage{booktabs}       
\usepackage{amsfonts}       
\usepackage{nicefrac}       
\usepackage{microtype}      
\usepackage{xcolor}         
\usepackage{amsmath}
\usepackage{graphicx}
\usepackage{amsthm}
\newtheorem{definition}{Definition}
\usepackage{float}      
\usepackage{placeins}   
\usepackage{amsfonts}       
\usepackage{wrapfig}        
\usepackage{amsthm}
\newtheorem{proposition}{Proposition}
\usepackage[font=footnotesize]{caption}
\usepackage{tabularx}
\title{Learning Reusable Behavioral Motifs for Continuous Animal Behavior Modeling}

\author{
    Jiyi Wang, Jingyang Ke, Bo Dai, Anqi Wu \\
  School of Computational Science and Engineering \\
  Georgia Institute of Technology, Atlanta, GA 30332 \\
  \texttt{\{jwang3617@, jingyang.ke@, bodai@cc., awu36@\}gatech.edu}
}

\begin{document}

\maketitle

\begin{abstract}
Natural behavior evolves continuously and often reuses movement structure across contexts, but existing behavioral segmentation methods typically assign a single discrete syllable at each time point or impose restrictive generative assumptions. We introduce Motif-based Continuous Dynamics (MCD), a computational framework that models behavior as time-varying combinations of reusable behavioral motifs. MCD learns motifs as latent behavior-level features from transition dynamics, through spectral representation learning within a reinforcement learning framework. MCD then infers policies that combine these motifs across time or context, through offline imitation learning. This formulation avoids hard behavioral boundaries while capturing continuity and compositionality. Across simulated gridworld and continuous-control tasks, mouse maze navigation and pose data, MCD identifies reusable motifs that explain task and temporal variation. MCD therefore provides a flexible and interpretable framework for computational behavior analysis. Code is available in \url{https://github.com/G1NO3/MCD}.


\end{abstract}

\setlength{\abovedisplayskip}{1pt}
\setlength{\abovedisplayshortskip}{1pt}
\setlength{\belowdisplayskip}{1pt}
\setlength{\belowdisplayshortskip}{1pt}
\setlength{\jot}{1pt}

\setlength{\floatsep}{1ex}

\section{Introduction}
A central goal in animal behavior research is to discover structured representations of behavior from large-scale recordings. Existing approaches often do so by identifying recurring behavioral syllables—such as grooming, running, or sniffing—and partitioning continuous trajectories into discrete units~\citep{wiltschko2015mapping, weinreb2024keypoint, luxem2022identifying, hsu2021b, berman2014mapping}. Despite their success in identifying structured behavioral patterns, these methods typically assign a single behavioral label to each time point and impose a fundamentally discrete view of behavior.

Two important properties of natural behavior are difficult to capture under this discrete formulation. First, \textbf{continuity}, real behavior evolves smoothly over time, whereas discrete segmentation introduces artificial boundaries and ambiguity during transitions. Second, \textbf{compositionality}, multiple movement tendencies may occur simultaneously, and related behaviors may reuse common components. For example, back and side grooming may share similar forelimb motions while differing in body orientation, and sniffing may co-occur with distinct locomotor patterns like walking or sitting. A single discrete label cannot directly represent such overlap. In addition, many approaches are either non-generative, such as clustering-based methods (\cite{hsu2021b, berman2014mapping}), or rely on restrictive assumptions like linear or Markovian dynamics (\cite{wiltschko2015mapping, weinreb2024keypoint, luxem2022identifying}), which may limit the behaviors they can represent or generate. These limitations motivate modeling behavior as a continuous control process rather than exclusively as a sequence of discrete states.

To address these limitations, we propose a framework for continuous behavior modeling based on reinforcement learning (RL). We view behavior as arising from an underlying policy and infer this policy from fixed behavioral trajectories through offline imitation learning. Rather than treating behavior as transitions between discrete syllables, our framework represents behavioral dynamics through continuously varying policies.

Motivated by the hypothesis that animals generate behavior by recombining a finite set of reusable components over time~\citep{santuz2019modular, flash2005motor}, we introduce \emph{Motif-based Continuous Dynamics discovery (MCD)}. Operationally, MCD defines a behavioral motif as a reusable, behavior-level latent state–action feature learned from observed transition dynamics. We use spectral representation learning within the RL framework~\citep{dai2017learning, ren2022spectral, shribak2024diffusion} to learn the shared motifs $\phi(s,a)$, and offline imitation learning to estimate coefficients $u$ that combine these motifs to parameterize the policy through $Q=\phi(s,a)^\top u$. The behavioral pattern captured by a motif could be low-level movement kinematics or high-level navigation strategies, depending on the state and action representations selected for the scientific question. For example, a free-moving mouse may show both face grooming and side grooming behaviors. We may infer common "grooming" motifs and different motifs involving different body parts. Our intended interpretation here is at the behavioral level, and connections to muscle synergies or neural motor primitives are presented in the Discussion.

\begin{wrapfigure}{R}{0.45\textwidth}
\vspace{-0.15in}
    \centering
    \includegraphics[width=0.45\textwidth]{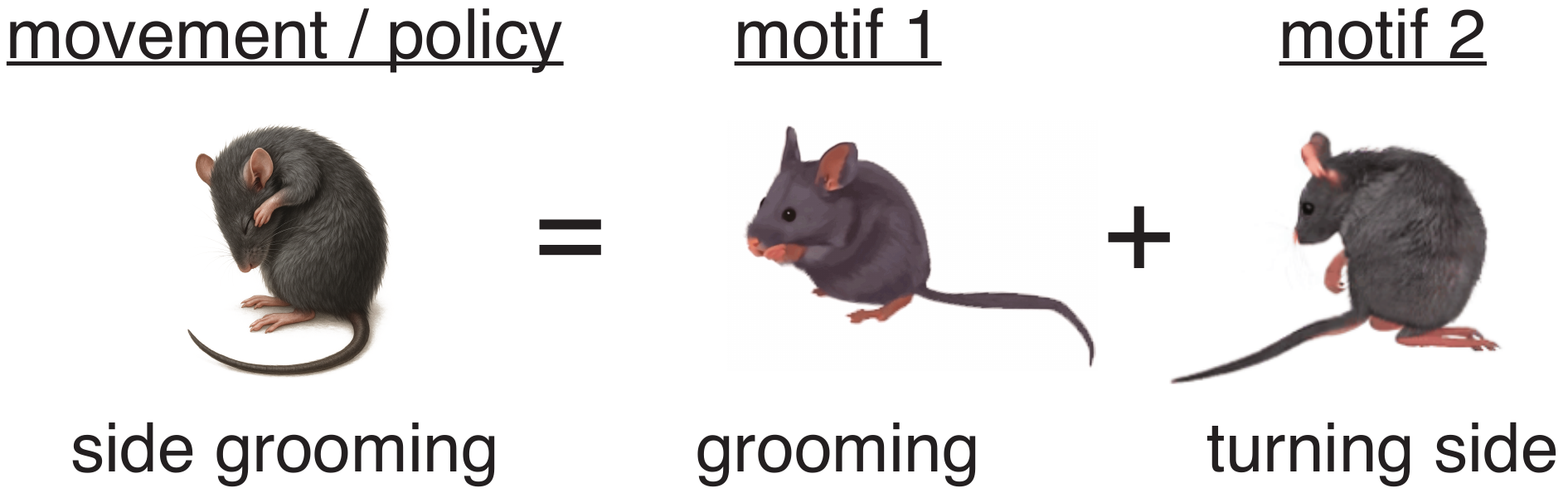} 
    \hspace{1mm}
\vspace{-0.2in}
\caption{\captionsize MCD models behavior as continuous compositions of reusable motor motifs. For example, ``side grooming'' may emerge from the combination of grooming and turning-side motifs.}
    \label{fig:overview}
\vspace{-0.1in}
\end{wrapfigure}

A key feature of MCD is that the policy is represented as a continuous combination of reusable motifs whose contributions can vary across tasks or evolve smoothly over time. Multiple motifs can be active simultaneously, allowing complex behaviors to emerge through dynamic composition rather than exclusive discrete switching. For example, side grooming may combine a "grooming" motif with a "turning" motif (Fig.~\ref{fig:overview}). Therefore, this formulation naturally captures continuity and compositionality.

Taken together, MCD provides a unified computational framework for modeling animal behavior as continuously evolving combinations of reusable behavioral motifs. Across simulated gridworld and continuous-control tasks, mouse maze navigation behavior, free-moving behavior, and macaque drawing behavior, MCD identifies reusable motifs that capture continuity and compositionality. These results support MCD as a flexible and interpretable framework for computational behavior analysis.

\section{Related work}\label{sec:related_work}

\paragraph{Animal behavior analysis.}
A major direction in animal behavior research is to discover reusable behavioral structure from large-scale recordings. Existing approaches typically do so by identifying recurring behavioral syllables, such as grooming or locomotion, using supervised classification (SIPEC~\citep{marks2022deep}, MARS~\citep{segalin2021mouse}), clustering-based methods (B-SOiD~\citep{hsu2021b}, MotionMapper~\citep{berman2014mapping}, SemiSeg~\citep{whiteway2021semi}), or hidden Markov model (HMM)-based approaches (VAME~\citep{luxem2022identifying}, MoSeq~\citep{wiltschko2015mapping}, Keypoint-MoSeq~\citep{weinreb2024keypoint}, WARHMM~\citep{costacurta2022distinguishing}). Among them, Keypoint-MoSeq~\citep{weinreb2024keypoint} is based on the autoregressive HMM (AR-HMM) method and has become a widely used modern framework in animal behavior analysis. These methods have been successful in uncovering stereotyped behavioral patterns and relating them to neural activity. However, they generally model behavior through discrete latent states, making it difficult to capture continuously evolving and compositional dynamics across contexts and timescales.

A closely related direction applies reinforcement-learning-based imitation learning to animal behavior analysis~\citep{aldarondo2024virtual}. Unlike our work, however, these approaches do not explicitly decompose behavior into compact and interpretable motor motifs.

A key methodological distinction from prior behavior modeling approaches lies in how behavioral dynamics are represented. Existing dynamics-based methods often model state transitions $P(s'|s)$ directly through simplified assumptions such as Gaussian or Markovian dynamics~\citep{wiltschko2015mapping,weinreb2024keypoint}. In contrast, MCD models behavior through policy-conditioned dynamics within an RL framework, where transitions emerge from the interaction between environment dynamics $P(s'|s,a)$ and behavior policies $\pi(a|s)$ as $P(s'|s) = \sum_a P(s'|s,a)\,\pi(a|s)$. This formulation provides a more flexible representation of behavior while avoiding strong assumptions on state transitions.

\paragraph{Reinforcement learning.}
Reinforcement learning has been widely used for skill discovery and hierarchical control in robotics and embodied agents~\citep{eysenbach2018diversity,sharma2020dynamics,ajay2020opal}. However, these approaches are primarily designed for control optimization rather than scientific analysis of natural behavior. In contrast, we use RL as a framework for discovering interpretable and reusable behavioral motifs from offline animal behavior data.

Our work is particularly related to offline skill discovery methods such as OPAL~\citep{ajay2020opal} and SPiRL~\citep{pertsch2021accelerating}, which learn latent skill representations from demonstrations. Unlike these approaches, MCD represents policies through generalized linear combinations of shared motifs, providing more interpretable behavioral representations. Additional RL-related work is provided in Appendix~\ref{appendix:rl_related_work}.

\section{Preliminaries}\label{sec:prelim}

\textbf{Markov Decision Processes (MDP).} 
To formalize motifs underlying behavior, we model behavioral trajectories within the framework of Markov Decision Processes (MDPs). An MDP is defined as a tuple $\mathcal{M} = (\mathcal{S}, \mathcal{A}, r, P, \rho, \gamma, H)$, where $\mathcal{S}$ denotes the state space (e.g., pose keypoints capturing the animal’s condition and environment), $\mathcal{A}$ the action space that describes feasible movements, $r:\mathcal{S}\times \mathcal{A} \to [0, 1]$ the reward function that encodes the immediate utility toward an internal goal, $P:\mathcal{S} \times \mathcal{A} \to \Delta(\mathcal{S})$ the transition kernel, with $\Delta(\mathcal{S})$ representing distributions over $\mathcal{S}$; $\rho \in \Delta(\mathcal{S})$ the initial state distribution, $\gamma \in (0, 1)$ the discount factor, and $H$ the horizon. A policy $\pi:\mathcal{S}\times [H] \to \Delta(\mathcal{A})$ specifies a distribution over actions given the current state. We assume that observed behavior is generated by a policy optimizing an intrinsic reward that reflects internal motivations. Under this view, behavioral trajectories arise from the interaction between the policy and environment dynamics.

Following standard notations, we define the value function $V(s) := \mathbb{E}\left[\sum_{t=0}^{H} \gamma^t r(s_t, a_t)|s_0 = s\right]$ and the action-value function $Q(s, a) = \mathbb{E}\left[\sum_{t=0}^{H} \gamma^t r(s_t, a_t)|s_0 = s, a_0 = a\right]$, the expected discounted cumulative rewards under policy $\pi$. These satisfy the Bellman equations:
\begin{align}\label{bellman}
    Q_h(s,a) = r(s,a)
      + \gamma\,\mathbb{E}_{s'\sim P(\cdot\mid s,a)}\bigl[V_{h+1}(s')\bigr], \quad
      V_h(s) = \mathbb{E}_{a\sim \pi(\cdot\mid s)}\bigl[Q_h(s,a)\bigr].
\end{align}



\textbf{Offline Imitation Learning.} We adopt an offline imitation learning (IL) setting to infer a policy $\pi$ that matches observed behavior. Instead of interacting with the environment, we are given a dataset of transitions collected from an expert (the animal): $\mathcal{D}=\left\{(s_i,a_i,s'_i)|(s,a)\sim\tau^e, s'\sim P(\cdot|s,a), i=1,2,...,N\right\}$, where $\tau^e$ denotes the empirical state–action distribution. The goal is to recover a policy consistent with this dataset.

\section{Methods}\label{sec:skil}

Given the MDP formulation, we frame motif discovery from a control-theoretic perspective. Motifs serve as fundamental building blocks that enable the construction of policies and reward functions, thereby providing a compact representation of behavioral dynamics.

\begin{definition}[Motif Set]\label{def:motif_set}
A transition kernel $P(s'|s, a)$ can be expressed via a spectral decomposition:
\begin{equation}\label{eq:p_linear}
    P(s'|s, a) = \phi(s, a)^\top \mu(s')q(s'),
\end{equation}
where $\phi: \mathcal{S} \times \mathcal{A} \rightarrow \mathbb{R}^d$, $\mu: \mathcal{S} \rightarrow \mathbb{R}^d$, and $q \in \Delta(\mathcal{S})$ is a parametrized probability distribution over the state space. We define $\phi$ as the \textbf{motif} set, and parameterize the reward as $r(s, a) = \phi(s, a)^\top w$.
\end{definition}
Spectral decomposition is a standard tool in RL representation learning \citep{ren2022spectral, shribak2024diffusion, zhang2022making}. Here, it provides a way to extract latent features directly from transition kernel, without imposing extra explicit assumptions. As a result, the learned motifs reflect intrinsic structure in the behavior data.

Given the motif definition, substituting Eq.~\ref{eq:p_linear} and the linear reward model into Eq.~\ref{bellman}, we get:
{\small\begin{align}\label{eq:q_linear}
Q(s,a) = r(s,a) + \gamma \!\!  \int\!  V(s')P(s' | s,a) \mathrm{d}s' = \phi(s,a)^\top\! \!\left[w + \gamma \! \! \int\!  V(s') \mu(s')q(s') \mathrm{d}s'\right]= \phi(s,a)^\top u,\!
\end{align}}
where $u=w + \gamma \int V(s')\,\mu(s')\,q(s')\,\mathrm{d}s'$. Thus, \( Q(s,a) \) can be expressed as a linear combination of motif features \( \phi(s,a) \), linking motifs to the policy. Following the maximum entropy RL framework~\citep{haarnoja2018soft}, the optimal policy takes the form:
\begin{align}\label{eq:softpolicy}
    \pi(a|s) &= \arg\max_{\pi}\; \mathbb{E}_{\pi}[Q(s, a)] + H(\pi) = \arg\max_{\pi} \frac{\exp(\phi(s, a)^\top u)}{\sum_{a' \in \mathcal{A}} \exp(\phi(s, a')^\top u)},
\end{align}
where $H(\pi) := -\sum_{a\in \mathcal{A}}\pi(a|s)\,\log\bigl(\pi(a|s)\bigr)$ is the entropy.

\begin{proposition}\label{prop1}
The policy in Eq.~\ref{eq:softpolicy} arises as the maximum-entropy solution induced by the spectral representation of the transition kernel. The motifs $\phi$ form a basis that can represent any max-entropy policy through an appropriate choice of $u$.
\end{proposition}

We interpret $\phi(s,a)$ as motifs because it provides a shared basis for transitions, policies, and rewards. Policies describe how behavior evolves, while rewards capture why it evolves. Since $\phi$ is derived solely from transition dynamics, it encodes intrinsic, reusable motor structure independent of specific tasks. The coefficients $u$ capture task-dependent modulation, yielding an interpretable decomposition of behavior as compositions of shared motifs. Unlike prior dynamics-based approaches that directly model $P(s'|s)$ under restrictive assumptions, our formulation represents behavior through a shared motif basis, avoiding explicit parametric dynamics models.

Together, Def.~\ref{def:motif_set} and Prop.~\ref{prop1} show that our framework captures behavioral dynamics without relying on restrictive model assumptions, in contrast to clustering-based methods (which lack dynamics) and traditional dynamics models (which impose strong assumptions). Next, we describe how to learn $\phi(s,a)$ and $\mu(s')$ (motif discovery) and $u$ (policy learning) from data. The learning procedure differs between discrete and continuous settings.



\subsection{Discrete Version}\label{sec:discrete_version}

\textbf{Motif discovery.} For discrete state-action spaces, motif representations can be learned via (1) classical spectral methods such as singular value decomposition (SVD)~\citep{golub1971singular, golub2013matrix, trefethen2022numerical}, or (2) spectral decomposition–based representation learning~\citep{ren2022spectral,haochen2021provable}. We adopt the latter and optimize 
\begin{align}\label{eq:discrete_motif}
\phi(s,a),\mu(s')=\arg\min_{\phi,\mu}\;\left\|P(s'|s,a)-\phi(s,a)^\top\mu(s')q(s')\right\|^2.
\end{align}
We set the auxiliary distribution to $q(s')=\rho(s')=\int \tau^e(s,a)\,da$, the marginalized state distribution. The learned motif set $\phi(s,a)$ is then used for policy learning.



\textbf{Motif-based policy learning.} We estimate the policy $\pi(a|s)$ (Eq.~\ref{eq:softpolicy}) via maximum likelihood estimation (MLE), i.e., by optimizing the following objective to solve for \( u \):
\begin{align}\label{MLEu}
\max_{u} \;\; \mathbb{E}_{(s,a)\sim\tau^e} \left[ \log \pi(a | s) \right] = \max_{u} \;\; \mathbb{E}_{(s,a)\sim\tau^e} \left[  \log \frac{\exp(\phi(s,a)^\top u)}{ \sum_{a'\in\mathcal{A}} \exp\left(\phi(s,a')^\top u\right) }\right].
\end{align}

\subsection{Continuous Version}\label{sec:continuous_version}

The continuous setting introduces two key challenges. First, the decomposition in Eq.~\ref{eq:p_linear} is too restrictive to capture realistic behavioral dynamics. For example, a common model assumes additive structure, $s' = h(s,a) + \epsilon$, where $h$ is a dynamics function and $\epsilon$ is Gaussian noise. This may not be well captured by the multiplicative form $\phi(s,a)^\top \mu(s')q(s')$, suggesting the need for a more flexible parameterization.

While learning from discrete data is relatively straightforward, the continuous setting introduces two key challenges. First, the decomposition in Eq.~\ref{eq:p_linear} is too restrictive to capture realistic behavioral dynamics. For example, consider a common and biologically plausible additive behavioral dynamics, $s' = h(s,a) + \epsilon$, where $h$ is a dynamics function and $\epsilon$ is Gaussian noise. This dynamics contrast with the multiplicative form \( \phi(s,a)^\top \mu(s')q(s') \), suggesting the need for a more flexible parameterization. Second, policy learning becomes intractable: the partition function in Eq.~\ref{MLEu} requires integration over a continuous action space with infinite actions that are impossible to enumerate. To address these challenges, we adopt an alternative formulation.


\textbf{Motif discovery.} We model the transition kernel $P(s'|s,a)$ as an energy-based model (EBM) \citep{shribak2024diffusion}:
{\small \begin{align}\label{eq:p_ebm}
    P(s'|s,a)= q(s')\exp\left(\psi(s,a)^\top\nu(s')- \log Z(s, a)\right),\quad Z(s, a)=\int q(s')\exp(\psi(s,a)^\top\nu(s')) ds',
\end{align}}
where $\psi:\mathcal{S}\times\mathcal{A}\to\mathbb{R}^g$ and $\nu:\mathcal{S}\to\mathbb{R}^g$ are neural-network feature maps and $Z(s,a)$ is the partition function. Compared to Eq.~\ref{eq:p_linear}, this EBM formulation yields more expressive motif representation and also produces more stable gradients during training.



\begin{proposition}[Connection to Motif Definition]
The EBM formulation induces a representation equivalent to Eq.~\ref{eq:p_linear}, where $\phi(s,a)$ and $\mu(s')$ are functions of $\psi$, $\nu$, and $Z(s,a)$. (Appendix \ref{sec:equivalence})
\end{proposition}



We learn $\psi$ and $\nu$ using noise-contrastive estimation (NCE) \citep{wu2018noise, ma2018noise, gutmann2010noise, gutmann2012noise}, 
which enables optimization of unnormalized statistical models without explicitly computing the partition function:
\begin{align}\label{eq:NCEP}
    \max_{\psi, \,\nu}\;\;\mathbb{E}_{\substack{(s,a) \sim \tau^e,\\ s'\sim P(\cdot|s,a),\ s''_i\sim\rho}} \left[\log \frac{F(s,a,s')}{F(s,a,s')+\Sigma_{i=1}^kF(s,a,s''_i)}\right],
\end{align}
where $F(s,a,s')=\exp(\psi(s,a)^\top \nu(s'))$, \( s' \) denotes a positive sample drawn from the transition distribution \( P(\cdot | s,a) \), and \( s''_i \) for \( i = 1, \ldots, k \) are negative samples from marginalized state distribution $\rho(s)=\int \tau^e(s,a)da$.



\textit{\underline{Connection to behavioral dynamics.}} 
With simple algebra, we obtain the quadratic potential function $P(s' | s,a)\propto q(s')\exp\left( \| \psi(s,a) \|^2 / 2 \right) \exp\left( -\| \psi(s,a) - \nu(s') \|^2 / 2 \right)\left( \| \nu(s') \|^2 / 2 \right)$ from Eq.~\ref{eq:p_ebm}. By assuming $\nu$ as an identity map, $Z(s,a)$ as a constant and $q(s')\propto\exp(-||s'||^2/2)$, Eq.~\ref{eq:p_ebm} reduces to a Gaussian transition model: $P(s' | s,a) = \frac{1}{Z} \exp\left( -\|s' - \psi(s,a)\|^2/2 \right)$, which matches standard assumptions in behavioral modeling discussed in Sec.~\ref{sec:related_work}. Thus, Eq.~\ref{eq:p_ebm} provides a more general and flexible framework that extends classical dynamics-based approaches. Moreover, in practice, using Eq.~\ref{eq:p_ebm} results in a unimodal distribution over \( s' \) which closely matches the observed behavior data, while Eq.~\ref{eq:p_linear} fails to do so. Thus, while the discrete formulation is theoretically valid in continuous domains, we adopt the parameterization in Eq.~\ref{eq:p_ebm} for continuous state and action spaces, as it more effectively supports the learning of motif representations underlying animal behavior.

\textbf{Motif-based policy learning.} Since $Z(s,a)$ is intractable, we approximate $\phi$ via a learned mapping $f:\psi\rightarrow\phi$. This yields \( Q(s,a) = f\bigl(\psi(s,a)\bigr)^\top u \) for later policy learning. However, direct MLE is infeasible due to the continuous action space. This brings us back to the challenge of estimating an unnormalized energy function, $\pi(a|s)\propto \exp(Q(s,a)) \propto \exp(f(\psi(s,a))^\top u)$. Thus, it's reasonable to apply NCE here again, 
\begin{align}\label{eq:NCEu}
    \max_{f,\, u}\;\;\mathbb{E}_{\substack{(s, a) \sim \tau^e,\\ (s_i', a'_i)\sim\tau^e}} \left[\log \frac{G(s,a)}{G(s,a)+\Sigma_{i=1}^kG(s,a'_i)}\right],
\end{align}
where $G(s,a)=\exp(f(\psi(s,a))^\top u)$, $\{s,a\}$ are positive samples and $\{s'_i, a'_i\}$ are negative samples. Identifiability discussions are provided in Appendix~\ref{appendix:identifiability}.

\subsection{Understanding Animal Behavior via Motifs}\label{sec:understanding_behavior}

In this section, we discuss how motifs \( \phi \) and weights \( u \) can be used to describe animal behavior trajectories, by allowing \( u \) to vary across tasks or time points and dynamically modulate each motif. We consider two scenarios: (1) discrete state-action spaces in a multi-task setting, and (2) continuous state-action spaces in a time-varying setting. In both settings, $u(t)$ can be viewed as a matrix whose columns correspond to task- or time-specific coefficients over a shared motif basis.

\textbf{Scenario 1.} Consider a mouse navigating a maze, where trajectories are discretized into a finite state and action space. We assume the animal switches among $T$ strategies, each associated with a distinct reward function. We first learn shared motifs $\phi(s,a)$ using Eq.~\ref{eq:discrete_motif}, and then fit task-specific policies $\pi(a|s,t)$ via Eq.~\ref{MLEu}. This yields $T$ sets of coefficients $u(t)$, one per task, all defined over a common motif basis. This formulation captures behavioral variation across tasks as changes in motif weights while preserving shared structure.

Scenario 1 mainly illustrates how to capture shared structure across multiple discrete behavioral strategies through compositional reuse of motifs. However, many natural behaviors are not organized as discrete task switches, but instead evolve continuously over time with gradual transitions and long-range temporal dependencies. To capture these properties, we next consider a continuous state-action setting with time-varying motif weights.

\textbf{Scenario 2.} We now consider a more naturalistic setting in which a mouse freely moves in a cage, with states represented by continuous body-part coordinates and actions represented by velocities. We learn $\psi(s,a)$ using Eq.~\ref{eq:NCEP}, and estimate $f$ and $u(t)$ via Eq.~\ref{eq:NCEu}. This results in motifs $\phi(s,a)=f(\psi(s,a))$, and policies constructed as time-varying combinations of these motifs through $u(t)$.

In this scenario, the policy is represented by a generalized linear combination of basis motifs (Eq.~\ref{eq:NCEu}), with two key properties: (1) \textbf{Continuity.} Unlike models with abrupt discrete switches, our method has motif weights that evolve smoothly over time, capturing gradual behavioral changes. (2) \textbf{Compositionality.} Unlike models assigning one discrete label to one timestep, our method has multiple motifs that can be active simultaneously, allowing behaviors to be expressed as combinations of shared components.


Taken together, MCD provides a flexible and interpretable representation of behavioral dynamics as continuous compositions of shared motifs. In Sec.~\ref{sec:exp}, we demonstrate (1) what motifs we have learned, and (2) how they are combined to generate behavior across simulated and real datasets.

\subsection{Reward Recovery}\label{sec:rewardrecovery}
Given the learned motif weights $u(t)$, we can further recover reward parameters \( w(t) \) in \( r(s,a,t) = \phi(s,a)^\top w(t) \) via 
\begin{align}
    w(t) = u(t) - \gamma \int V(s', t)\, \mu(s')\, q(s')\,\mathrm{d}s',
\end{align}
where \( V(s, t) = \log \sum_a \exp Q(s,a,t) \). This allows us to recover the task-specific or time-varying internal reward function $r(s,a,t)$ used by animals. This formulation aligns with inverse reinforcement learning (IRL)~\citep{ziebart2008maximum}. In the context of animal behavior~\citep{ke2025inverse, zhu2023multi, ashwood2022dynamic}, recovering such rewards offers insight into internal driving factors for decision-making processes. Since $V(s,t)$ is only tractable in discrete settings, we validate this approach by visualizing inferred rewards in the discrete experiments.



\section{Experiments}\label{sec:exp}

We organize the experiments as a progression from controlled validation to real continuous animal behavior. We first study simulated environments with known reward structure, including a multi-task gridworld and a continuous Lunar Lander setting (Appendix~\ref{appendix:lunarlander}), to evaluate whether MCD can recover interpretable motifs and reward structure. We then apply MCD to real mouse behavior, first in maze navigation and finally in freely moving natural behavior. Together, these experiments evaluate the practical applicability of MCD across discrete and continuous domains.

\subsection{Application to simulated multi-task gridworld}\label{sec:exp1}

\begin{figure}{}
    \centering
    \includegraphics[width=\textwidth]{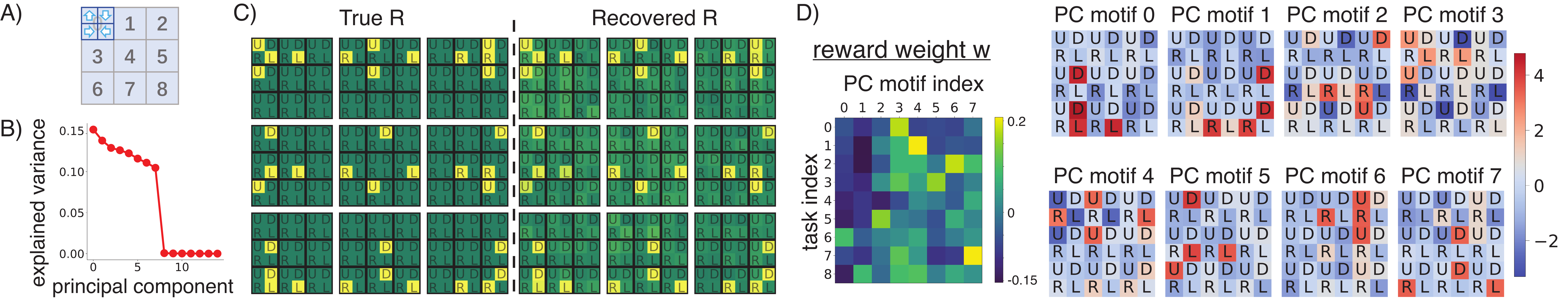}
  \caption{\textbf{A.} State-action map: each grid contains four action values (Up, Down, Left, Right). In task $i$, rewards are assigned to actions moving toward the $i$th location. \textbf{B.} Explained variance of the top 15 principal components; variance drops near zero after PC7. \textbf{C.} Left: ground-truth rewards for all 9 tasks (yellow indicates high reward and green indicates low reward). Right: recovered rewards. \textbf{D.} Task-specific reward weights $w$ and the top 8 principal-component motifs derived from $\phi$.}
    \vspace{-0.2in}
  \label{fig:exp1}
\end{figure}


The environment consists of a $3\times3$ gridworld with four actions per state: Up, Down, Left, and Right (Fig.~\ref{fig:exp1}A). In task $t\in\{0,\dots,8\}$, rewards are assigned to $(s, a)$ pairs that move toward target state $t$. Fig.~\ref{fig:exp1}C (left) shows the ground-truth reward for all tasks. For each episode, the agent starts from a random state and navigates toward the target. Additional details are provided in Appendix~\ref{sec:app_gridworld}.

Following Sec.~\ref{sec:discrete_version}, we learn a shared motif representation $\phi(s,a)$ and use it to parameterize task-specific policies $\pi(a|s,t)$. Since computational complexity scales linearly with the motif dimension, we use a relatively large representation size ($d=64$) to maximize coverage of the motif space. Using the learned representation, we recover the reward function $r(s,a,t)=\phi(s,a)^\top w(t),$ as described in Sec.~\ref{sec:rewardrecovery}. The recovered rewards closely match the ground truth (Fig.~\ref{fig:exp1}C, right), achieving a Pearson correlation of 0.96. This suggests that the learned motifs are sufficient to reconstruct task structure from behavior alone. Visualizations of these motifs are shown in the Appendix~\ref{sec:app_gridworld}. Discussion of motif dimension is provided in Appendix~\ref{app:ablation_motif_dimension}.

To interpret the learned representation, we apply principal component analysis (PCA) to the motif matrix $\phi$. Only eight principal components explain most of the variance (Fig.~\ref{fig:exp1}B), indicating that the effective motif space is low-dimensional. We visualize the top principal-component motifs and their corresponding task-specific coefficients $w(t)$ in Fig.~\ref{fig:exp1}D. For example, motif 0 assigns positive values to transitions toward the bottom-left region and negative values to transitions toward the middle-right region, corresponding to movement from State $5$ toward the State $6$. Consistent with this interpretation, the corresponding motif weight in $w$ is strongly positive for Task 6 and negative for Task 5. Similar structure appears across other motifs and tasks, demonstrating that the learned motifs provide interpretable building blocks for reward composition. We further evaluate MCD in a simulated continuous-control task Lunar Lander in Appendix~\ref{appendix:lunarlander}, demonstrating that motif discovery can generalize to the continuous case.

\subsection{Application to animal navigation behavior}\label{sec:exp2}



\begin{figure}{}
    \centering
    \includegraphics[width=\textwidth]{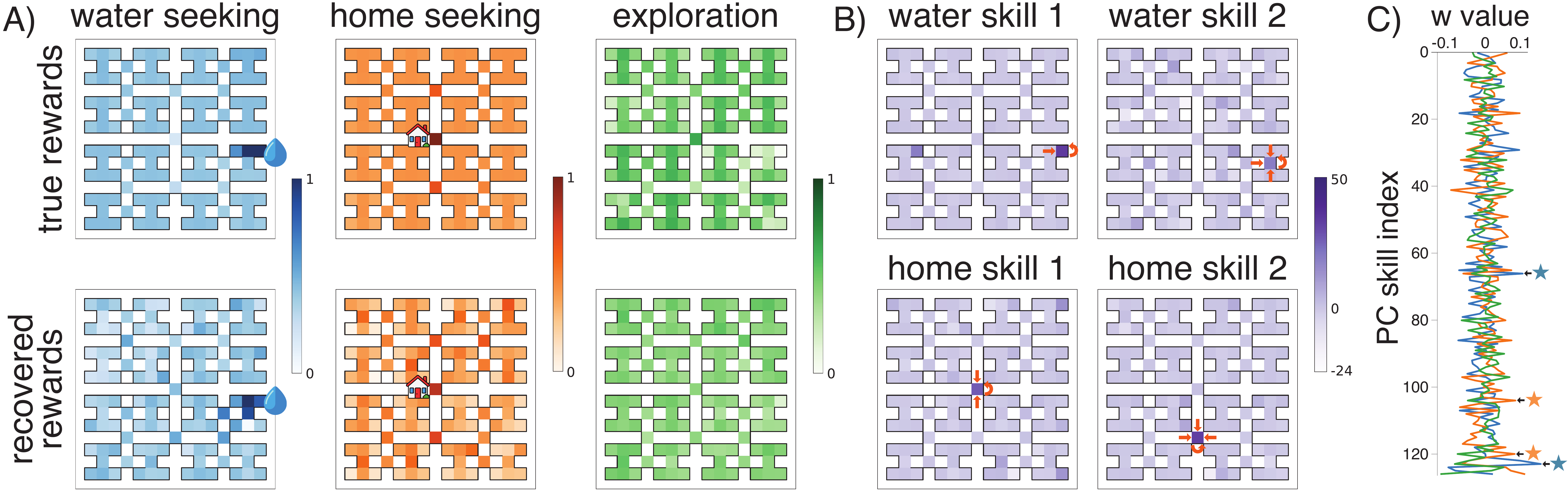}
    \vspace{-0.1in}
  \caption{The maze contains 63 T-junctions and 64 terminal nodes. The central root is connected to the home cage. The water port is placed at one terminal node. At each T-junction, an action denotes a transition to one of the adjacent nodes (the parent or the left/right child). The house and droplet icons indicate the home cage and the water port, respectively. \textbf{A.} Ground-truth and recovered rewards for the three behavioral modes. Darker colors denote higher reward. \textbf{B.} Dominant motifs for the water-seeking and home-seeking behaviors. Orange arrows denote high-value motifs. Darker colors denote higher state values $v(s)$ defined in the main text. \textbf{C.} Task-specific coefficients $w$ for all motifs (blue, orange, and green indicate water seeking, home seeking, and exploration, following \textbf{A}). Positive coefficients $w$ indicate a motif contributes highly to that recovered reward. The two blue stars highlight the two water-seeking motifs in \textbf{B}, and orange stars highlight home-seeking motifs in \textbf{B}.}
    \vspace{-0.3in}
  \label{fig:exp2}
\end{figure}

\textbf{Dataset and model setup.} We evaluate our method on a real discrete animal behavior dataset from \cite{rosenberg2021mice}. In this experiment, a thirsty mouse is trained to start from home and navigate to the water port in a binary-tree maze (Fig.~\ref{fig:exp2}A). The states correspond to the mouse’s location in the maze. The actions correspond to transitions to the left/right parent or child node. Although the dominant behavioral drive is water seeking, the mouse also explores unfamiliar regions and returns to the home cage for shelter. These competing motivations cannot be adequately captured by a single reward function, making this dataset well-suited for studying reusable motifs underlying multiple behavioral strategies.

We first apply the segmentation algorithm from \cite{ke2025inverse} to partition trajectories into three interpretable behavioral modes: water seeking, home seeking, and exploration, and apply MCD to infer the corresponding reward functions (Fig.~\ref{fig:exp2}A, top row). Water seeking denotes directed movement toward the water port, home seeking toward the maze exit, and exploration contains the remaining non-directed trajectories. While the true internal rewards are unknown, these inferred rewards generate policies that closely match observed behavior and therefore serve as effective ground truth. Using the discrete formulation in Sec.~\ref{sec:discrete_version}, we learn a shared motif representation across all three tasks. We set the motif dimension to $d=127$ to maximize coverage of the motif space (Appendix~\ref{app:ablation_motif_dimension}).

\textbf{Results.} The recovered task-specific rewards capture the most prominent part of inferred ground truth (Fig.~\ref{fig:exp2}A, bottom row). In the water-seeking condition, rewards (dark areas) peak near the water port and along trajectories leading toward it. In the home-seeking condition, the dominant reward concentrates around the home cage. Exploration produces a comparatively uniform reward landscape, with reduced value near the water port, suggesting temporary suppression of water-seeking behavior. We also construct a "perfect" reward map where (1) the water-seeking reward function is one-hot at the water port, (2) the home-seeking reward function is one-hot at the home cage, and (3) the exploration reward function has a uniform small reward everywhere. We calculate the correlation of our recovered map and ground truth with the perfect map, and find their correlations are similar ($0.4$). Therefore, our recovered reward map matches the prominent part of the ground truth.

To interpret the learned representation, we apply PCA to the motif matrix and visualize representative principal-component motifs. Fig.~\ref{fig:exp2}B shows selected motifs with the largest task-specific coefficients $w$ (top row: water-seeking, bottom row: home-seeking) in Fig.~\ref{fig:exp2}C. For each selected motif $\phi_i(s,a)$, we scan the whole state-action space and pick those $(s,a)$ pairs with the highest motif values (shown with orange arrows). Purple shading indicates the state value $v_i(s')$, which is the average motif value pointing to this state: $v_i(s')=\sum_{\{(s,a)|P(s'|s,a)=1\}} \phi_i(s,a)$. 
 
Water-related motifs promote transitions toward the water port, whereas home-related motifs favor movement toward the cage. Interestingly, most motifs are dominated by a single behavioral mode rather than shared uniformly across tasks (Fig.~\ref{fig:exp2}C), reflecting competition between underlying motivations. Details for these motifs are provided in Appendix~\ref{sec:app_labyrinth}. Unlike previous reward-discovery approaches on this dataset that lack motifs~\citep{ashwood2022dynamic,ke2025inverse}, MCD learns a shared motif basis that simultaneously explains multiple behavioral strategies through different motif combinations.


\subsection{Application to animal free-moving behavior}\label{sec:exp3}

\begin{figure*}[t]
    \centering
    \includegraphics[width=0.95\textwidth]{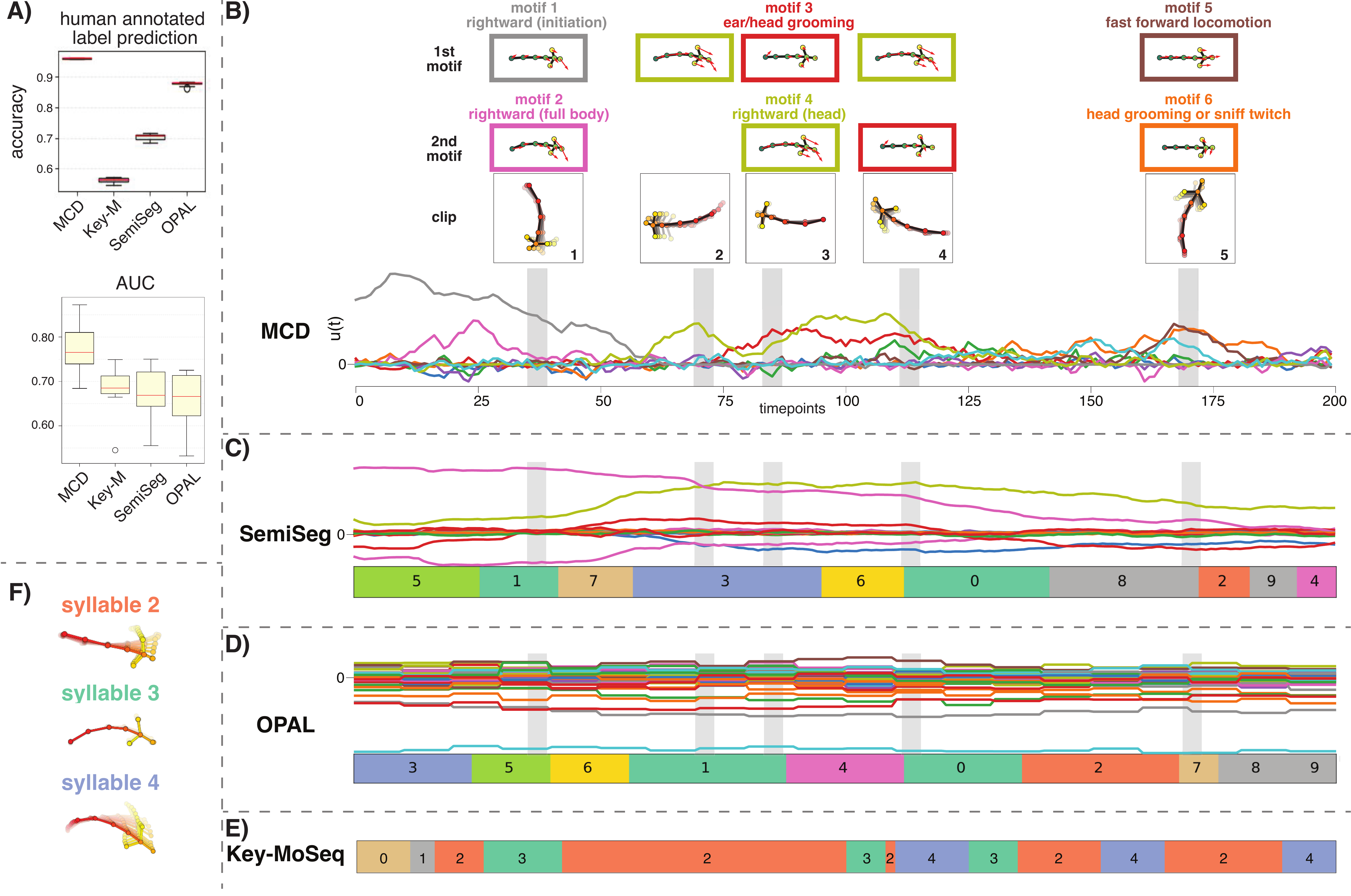}
      \caption{\textbf{A.} Boxplot for human-annotated label prediction accuracy (top) and trajectory discrimination AUC (bottom). AUC is evaluated on 10 sequences (length=100) with 1000 negative samples per sequence. \textbf{B.} Example motif weights $u(t)$ and representative motifs inferred by MCD. The middle "clip" row shows real animal behavior clips. \textbf{C-D.} Latent representations and segmentation results for SemiSeg and OPAL. \textbf{E-F.} Segmentation results and representative syllables from Keypoint-MoSeq~\citep{weinreb2024keypoint}.}
        \vspace{-0.3in}
      \label{fig:exp3}
\end{figure*}

\textbf{Dataset and model setup.} To evaluate MCD on continuous natural behavior, we use a free-moving mouse dataset from \cite{weinreb2024keypoint}. States correspond to the $(x,y)$ coordinates of eight body parts, including the head, nose, ears, and spine nodes. Actions are defined as body-part velocities, $a_t = (s_{t+1}-s_t)/\delta t,$ with $\delta t=1$ ($1/30$s in real time). This dataset contains rich mixtures of grooming, locomotion, and postural adjustment behaviors, making it suitable for testing whether complex movements can be represented through a compact set of reusable motifs.


We use the continuous formulation in Sec.~\ref{sec:continuous_version} to learn motif representations $\phi(s,a)$ and time-varying weights $u(t)$ (Sec.~\ref{sec:understanding_behavior}). To encourage temporal smoothness, we place a Gaussian random-walk prior on $u(t)$: $u(t)\sim\mathcal{N}(u(t-1),\sigma^2 I).$ We optimize $f$ and $u(t)$ via coordinate descent. Since our goal is to analyze behavioral structure rather than recover rewards, we focus on interpreting $\phi$ and $u(t)$ rather than performing IRL. Performance saturates beyond motif dimension $d=64$ (Appendix~\ref{app:ablation_motif_dimension}), so we use $d=64$ in practice. Additional training details are provided in Appendix~\ref{sec:app_keymoseq}.

We compare MCD against three representative baselines (Sec.~\ref{sec:related_work}): Keypoint-MoSeq~\citep{weinreb2024keypoint}, an HMM-based behavior segmentation method; SemiSeg~\citep{whiteway2021semi}, a clustering-based approach; and OPAL~\citep{ajay2020opal}, an RL-based skill learning framework.



\textbf{Results.} To evaluate whether the inferred latent weights $u(t)$ capture meaningful behavioral structure, we train a two-layer neural network to decode human-annotated labels. MCD achieves the highest test accuracy (Fig.~\ref{fig:exp3}A top, $p<0.05$), suggesting that its motif representations align closely with behavioral dynamics. Additional details are provided in Appendix~\ref{sec:appendix_human_label}.

We further evaluate each model's ability to distinguish real trajectories from mismatched state-action pairs using the area under the receiver operating characteristic curve (AUC). This method allows direct comparison between our unnormalized energy function and other models' normalized likelihood. Given $(s,a),(s',a')\sim\tau^e$, positive samples are true pairs $(s,a)$, while negative samples are mismatch pairs $(s,a')$. With respect to the prediction score for AUC, we use action log-likelihood for Keypoint-MoSeq and OPAL, Gaussian log-likelihood for SemiSeg, and negative energy $\phi(s,a)^\top u(t)$ for MCD. MCD achieves the highest AUC across all baselines (paired t-test, $p<0.05$; Fig.~\ref{fig:exp3}A bottom), indicating the strongest ability to capture real dynamics.


Beyond quantitative evaluation, we also qualitatively visualize the learned motifs and their temporal activations on an example behavior sequence of length=250 (Fig.~\ref{fig:exp3}B bottom). We extract five short clips ((Fig.~\ref{fig:exp3} B, the third row), each associated with a distinct motif combination (Fig.~\ref{fig:exp3} B top two rows). We visualize the dominant motifs for each clip, and show their average poses and actions, computed by averaging $(s,a)$ that most strongly activate that motif. Additional visualizations are provided in Appendix~\ref{sec:app_keymoseq}. For comparison, we visualize latent trajectories from SemiSeg and OPAL (Fig.~\ref{fig:exp3}C-D), together with Keypoint-MoSeq segmentations (Fig.~\ref{fig:exp3}E). The segmentations of SemiSeg and OPAL are calculated through KMeans ($k=10$) on their first 10 PCs of the latents.

MCD captures both transitions and mixtures of behaviors through smoothly varying motif activations. In Clip 1, right-turning behavior is represented by the joint activation of two turning motifs, with motif 1 emphasizing stronger head motion. Clips 2--4 exhibit alternating turning and grooming dynamics, reflected by alternating motif dominance of motif 3 and motif 4. Clip 5 combines forward locomotion and sniffing, captured through simultaneous activation of two motifs. Notably, motif 4 reappears across multiple clips, indicating that motifs are reusable components rather than clip-specific states.

In contrast, baseline methods show less consistent structure. For example, in Clip 2-4, SemiSeg assigns nearly identical latent states to qualitatively different behaviors. OPAL fails to produce interpretable and dominant motifs for interpretation. Both models lack recurring motifs, suggesting their latents can not decompose behaviors into a simple subset. Keypoint-MoSeq often assigns distinct behaviors to the same syllable. For example, Clip 2 and 3 receive the same label despite corresponding to turning and grooming behaviors respectively. Moreover, mixed behaviors in Clip 5 with simultaneous locomotion and sniffing are not represented compositionally in Keypoint-MoSeq.

Taken together, these results show that MCD captures animal behavior through a compact and reusable set of motifs whose continuous combinations explain complex behavioral dynamics. Additional ablations on motif dimension, policy noise distribution, dataset partitioning,  Gaussian random walk priors and temporal resolution are provided in Appendix~\ref{appendix:hyperparameter}. We further assess generative quality through rollout experiments in Appendix~\ref{appendix:rollout}. To show that MCD achieves more than just a smoothed discrete model, we carry out an ablation study in Appendix~\ref{appendix:motif-continuity}.

\subsection{Application to macaque character drawing}\label{sec:exp4}

\begin{figure}[htbp]
\vspace{-0.1in}
\centering
\includegraphics[width=0.8\linewidth]{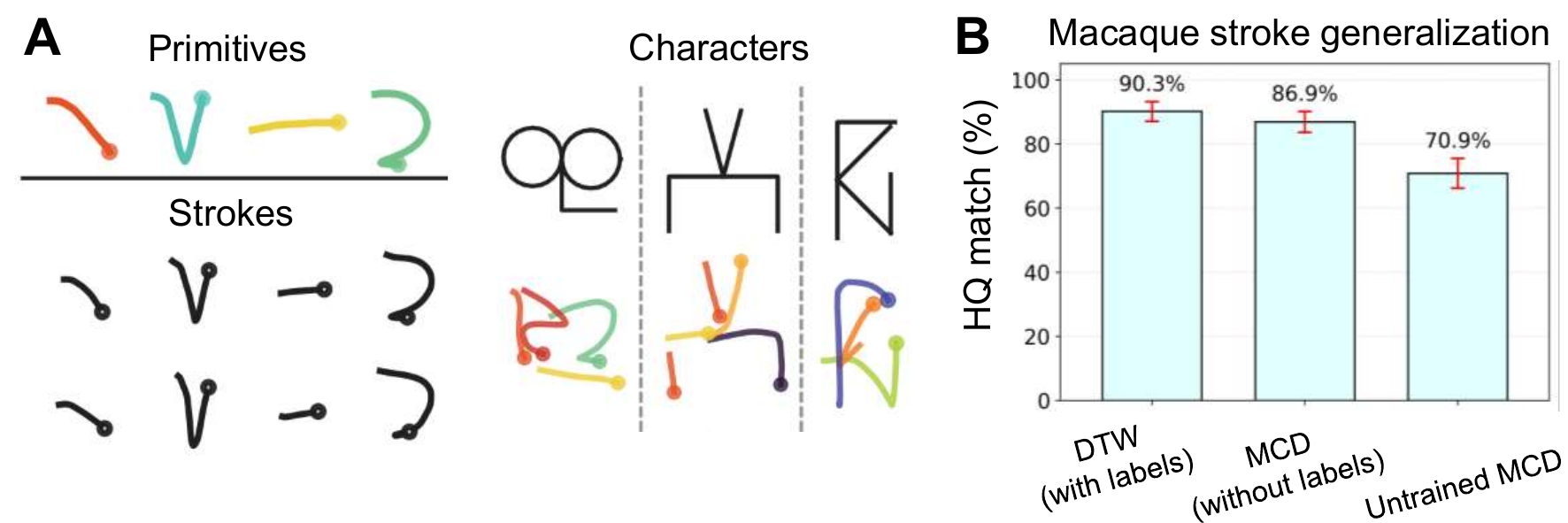}
\caption{\textbf{A.} The strokes can be categorized into different primitives, and the primitives are reused in different characters. \textbf{B.} Primitive HQ coverage for the test set. Bars show HQ rates.
Red error bars show $95\%$ bootstrap intervals. All methods use the same evaluation strokes.}
\label{fig:macaque-hq}
\vspace{-0.1in}
\end{figure}

We evaluate MCD on the macaque drawing dataset of
Tian et al.~\cite{tian2026neural}, in which monkeys
copy geometric figures on a touchscreen. After practising
simple shapes, the animals were presented with novel,
more complex figures. The original study found that
they reused familiar drawing strokes, or
\emph{primitives}, when producing these novel characters (Fig.~\ref{fig:macaque-hq}A).
We investigate
whether MCD can identify the same reusable primitives
in previously unseen drawings. We train MCD on isolated
primitives and practised characters, and evaluate it on
held-out novel characters.

Here, the state is defined as the 2D drawing position, and the action is the velocity. Different tasks correspond to different characters. Since each character consists of multiple strokes, and each stroke is a sequence of motifs, we first represent
each stroke $r$ by averaging its motif
features $\phi(s_{r,t},a_{r,t})$ over its $T_r$
time steps: $z_r = \frac{1}{T_r}\sum_{t=1}^{T_r}
\phi(s_{r,t},a_{r,t})$. We then run K-means on isolated primitives to obtain $K=19$ clusters and their centroids. The discovered clusters achieve $89.8\%$ Hungarian alignment (chance level=$14.7\%$) with the predefined labels.

To quantify primitive reuse in the novel characters, we measure whether these strokes in the novel characters can match the clusters constructed from the training dataset. Each test stroke representation
$z_r$ is compared with all 19 primitive centroids using cosine-based similarity $S(z_r,\mu_c) =1/\left(2-\cos\langle z_r,\mu_c\rangle\right)$ and assigned to its nearest primitive,
$\hat c_r=\arg\max_c S(z_r,\mu_c)$. The stroke is considered a high-quality (HQ) match if the similarity surpasses the threshold $\tau_c$. We set $\tau_c$ to the
2.5th percentile of the similarities within
each assigned group (Appendix~\ref{sec:app_macaque_drawing}). The HQ rate is then computed by averaging the
fraction of HQ strokes within characters in test dataset. MCD achieves $86.9\%$ without labels, close to the $90.3\%$ achieved by the original label-based
dynamic time warping (DTW) reference and
substantially above untrained MCD (Fig.~\ref{fig:macaque-hq}B).
These results indicate that MCD can discover useful primitives
and recognize their reuse in novel character compositions.

\section{Discussion}\label{sec:discuss}

In this work, we introduced \emph{Motif-based Continuous Dynamics discovery (MCD)}, a reinforcement-learning-based framework for understanding animal behavior through reusable motor motifs and their continuous compositions. Unlike traditional behavior segmentation approaches that model behavior as transitions between discrete syllables, MCD represents behavior as a combination of shared motifs learned directly from transition dynamics. Across simulated and real animal behavior datasets, we showed that MCD can recover interpretable motif structures, reconstruct reward functions, and provide more faithful representations of continuous pose dynamics than existing segmentation methods. Together, these results suggest that natural behavior is better characterized as continuous compositions of reusable motifs rather than abrupt switches between discrete behavioral states.

Our framework is motivated by longstanding ideas in neuroscience and motor control that complex movements arise from combinations of a finite set of reusable motor primitives~\cite{flash2005motor,bizzi1995modular,d2003combinations,tresch1999construction,hart2010neural}. MCD operationalizes this perspective within a reinforcement learning framework by treating motifs as low-level basis components shared across policies and rewards. Importantly, the learned motifs emerge directly from behavioral dynamics without requiring neural recordings, reward annotations, or hand-designed behavioral labels.

MCD also provides a principled framework for relating continuous behavioral dynamics to low-dimensional neural representations in future studies. The learned motifs $\phi(s,a)$ define low-level movement components that may serve as behavioral regressors for testing whether neural populations encode specific motor primitives. Meanwhile, the time (task)-varying policy weights and reward-related coefficients characterize how motifs are combined and modulated over time (task), potentially relating to control- and value-related neural signals.

Beyond neuroscience, MCD is also relevant for studying naturalistic behavior in ethology. By allowing motifs to co-occur and evolve continuously, the framework naturally captures multi-scale behavioral organization (Appendix~\ref{appendix:long-term}) without imposing discrete state assumptions. This makes MCD suitable for studying exploratory behavior, foraging, grooming, and other unconstrained behaviors that unfold across multiple timescales.

Several limitations remain for future work. First, the quality of the learned motifs depends on the quality of behavioral recordings, and tracking noise or pose estimation errors may degrade performance. Second, while the current framework focuses solely on behavior, incorporating multimodal information such as neural activity, sensory inputs, or environmental context may further improve motif discovery and interpretability. Finally, although MCD provides interpretable behavioral motifs, establishing direct correspondence between these motifs and specific neural dynamics still requires experimental validation.



\section*{Acknowledgement}\label{sec:acknowledgement}
This work was supported by NSF CAREER Award IIS-2540614.

\bibliography{reference}
\bibliographystyle{unsrt}

\newpage

\appendix

\section{Impact statement}\label{sec:impact_statement}

Beyond advancing animal behavior research, MCD has broader implications. Positively, a better understanding of motor control mechanisms could, for instance, inform new treatments for movement disorders or inspire more adaptable AI. On the other side, extending these principles to model human behavior carries ethical risks, such as perpetuating or amplifying societal biases present in training data. A robust ethical framework is essential to mitigate such risks in the development and application of these technologies.

\section{Reproducibility statement}\label{sec:repro}

We have taken several steps to ensure the reproducibility of our results. All source code for model training and evaluation is included in the supplementary material, allowing independent verification and replication of our experiments. The complete set of hyperparameter values is documented in the appendix. Additionally, the data preprocessing procedures and evaluation protocols are described in the main text. These resources provide sufficient information for reproducing the results reported in this paper.

\section{LLM usage}
In preparing this manuscript, we employed a large language model (OpenAI ChatGPT, GPT-5) as a writing assistant. The model was used exclusively for polishing English grammar, improving clarity, and suggesting more natural phrasing in certain sections of the text. All scientific content, experimental design, analyses, and interpretations were conceived, written, and verified by the authors. The LLM was not used to generate original research ideas, analyses, or results. To ensure accuracy, all model-suggested edits were carefully reviewed and, where necessary, modified by the authors.

\section{Licenses}
\label{sec:appendix_license}

All datasets, environments, and baseline implementations used in this work are publicly available and are properly credited through citations in the main text and references. We use these assets in accordance with their respective licenses and terms of use.

\paragraph{Gymnasium / LunarLander environment.}
The LunarLander environment from Gymnasium/OpenAI Gym is distributed under the MIT License, with copyright attributed to OpenAI and the Farama Foundation. Our usage complies with the corresponding open-source license terms. The official project page is:
\url{https://github.com/Farama-Foundation/Gymnasium}

\paragraph{Labyrinth navigation dataset.}
The labyrinth navigation dataset from~\citep{rosenberg2021mice} is released under the MIT License. We use the dataset and associated software in accordance with the license terms. The official project page is:
\url{https://github.com/markusmeister/Rosenberg-2021-Repository}

\paragraph{Keypoint-MoSeq.}
The Keypoint-MoSeq software and dataset~\citep{weinreb2024keypoint} are distributed under the Harvard University Non-Commercial Research and Academic Use Software License, which permits non-commercial academic research use. Our usage complies with these terms. The official project page is:
\url{https://github.com/dattalab/keypoint-moseq}.

\paragraph{Released assets.}
The implementation code released with this paper is provided solely for academic research and reproducibility purposes.

\section{Computation resources and execution time}

\paragraph{Hardware.}
All experiments were conducted on a single workstation equipped with an AMD Ryzen 7 7840H CPU, 16\,GB RAM, and an NVIDIA GeForce RTX 4060 Laptop GPU (8\,GB VRAM). The GPU driver version is 566.14 with CUDA 12.7 support.

\paragraph{Software.}
All models were implemented in PyTorch with standard CUDA acceleration. Detailed environment and dependency configurations are provided in the supplementary material.

\paragraph{Training cost.}
The computational cost varies across tasks:
\begin{itemize}
    \item Gridworld experiments: $\sim$10 minutes per run
    \item Labyrinth navigation: $\sim$1 hour per run
    \item Continuous animal behavior dataset: $\sim$8 hours per run
\end{itemize}

\paragraph{Reproducibility and scalability.}
All models are trained from scratch without pretraining. The experiments can be reproduced on a single consumer-grade GPU with comparable memory (e.g., 8\,GB VRAM). Additional exploratory or preliminary experiments not included in the paper required modest additional compute but follow similar resource profiles.\label{sec:appendix_compute_resource}

\clearpage
\section{Hyperparameter setting}
We train MCD using the following hyperparameters: 
\paragraph{General hyperparameters.}  
\begin{itemize}
  \item Discount factor: $\gamma = 0.99$
  \item Number of gradient update steps: $1 \times 10^6$
  \item Batch size: $256$
\end{itemize}

\paragraph{Motif representations.}  
The motif representation $\phi(s,a) \in \mathbb{R}^d$ and $\mu(s') \in \mathbb{R}^d$ were adopted with different motif dimensions $d$ depending on the task:
\begin{itemize}
  \item Gridworld: $d=64$
  \item Animal navigation: $d=127$
  \item Animal free-moving: $d=64$
\end{itemize}

\paragraph{Model architectures.}  
\begin{itemize}
  \item \textbf{Discrete version:} $\phi$ and $\mu$ are parameterized by one-hidden-layer neural networks (hidden size $=512$).
  \item \textbf{Continuous version:} 
  
    \begin{itemize}
      \item $f$: no hidden layer
      \item $\nu$: one hidden layer (hidden size $=512$), with one normalization layer to ensure the L2-Norm of the output is 1.
      \item $\psi$: no hidden layer, with one normalization layer to ensure the L2-Norm of the output is 1.
    \end{itemize}
    \item In both cases, $u$ and $w$ are just two matrices, with each column corresponding to $u(t)$ or $w(t)$ at a specific task or timepoint.
    \item Activation function: For all networks, if it has at least one hidden layer, then all of its activation functions are Exponential Linear Unit (ELU).

\end{itemize}

\paragraph{Learning rates.}  
\begin{itemize}
  \item \textbf{Discrete version:} 
  \[
  \phi: 1\times10^{-3}, 
  \mu: 1\times10^{-3}, 
  u: 3\times10^{-4}, 
  w: 3\times10^{-4}.
  \]
  \item \textbf{Continuous version:} 
  \[
  \psi: 5\times10^{-4}, 
  \nu: 5\times10^{-4},
  f: 3\times10^{-4}, 
  u: 3\times10^{-4}.
  \]
  During testing, $u$ and $f$ are further optimized on the new sequence using gradient descent with learning rate $1\times10^{-3}$.
\end{itemize}

We train SemiSeg and OPAL using the following hyperparameters:
\begin{itemize}
  \item Discount factor: $\gamma = 0.99$
  \item Number of epochs: $1 \times 10^6$
  \item Batch size: $4\times 250$ (4 sequences, each of length=250 because this is an RNN-based inference model)
  \item Latent dimension: $d=64$
  \item Learning rate: $1\times 10^{-4}$ (tuned to get better results)
\end{itemize}

Besides, the following loss coefficients are shared across three models for interpretability results.
\begin{itemize}
    \item Temporal smoothness Gaussian random walk loss: $10$
    \item Sparsity L1-loss: $0.1$
\end{itemize}

\label{appendix:hyperparameter}

\clearpage
\section{Approximating energy-based formulation with low-rank spectral decomposition}\label{sec:equivalence}


In this part, we show in detail the connection between the EBM formulation (Eq.~\ref{eq:p_ebm}) and the low-rank spectral decomposition formulation (Eq.~\ref{eq:p_linear}) of the transition kernel $P(s'|s,a)$.

From Eq.~\ref{eq:p_ebm}, by simple algebra, we obtain the quadratic potential function,

\begin{align}
P(s' | s,a) 
\propto &q(s')\cdot\exp\left( \| \psi(s,a) \|^2 / 2 \right)\cdot\exp\left( \| \nu(s') \|^2 / 2 \right) \nonumber\\
   &\cdot\exp\left( -\| \psi(s,a) - \nu(s') \|^2 / 2 \right). \label{eq:gaussian_kernel}
\end{align}

The term $\exp\left( -\frac{\| \psi(s,a) - \nu(s') \|^2}{2} \right)$ is the Gaussian kernel, for which we apply the random Fourier feature~\citep{dai2017learning, rahimi2007random} and obtain the spectral decomposition of Eq.~\ref{eq:p_ebm} as 
\begin{align}
 P(s' | s,a) = \langle \phi_\omega(s,a), \mu_\omega(s')q(s') \rangle_{\mathcal{N}(\omega)}, \label{eq:P_rff}
\end{align}

where \( \omega \sim \mathcal{N}(0, I) \) is the frequency in the Fourier domain, and
\begin{align}
\phi_\omega(s,a) = &\exp\left( -i \omega^\top \psi(s,a) \right) \nonumber\\
&\cdot\exp\left( \| \psi(s,a) \|^2 / 2 - \log Z(s,a) \right), \label{eq:phi_omega} \\
\mu_\omega(s') = &\exp\left( -i \omega^\top \nu(s') \right) \cdot\exp\left( \| \nu(s') \|^2 / 2 \right). \label{eq:mu_omega}
\end{align}

Note that Eq.~\ref{eq:P_rff} needs infinite $\omega$ to calculate the expectation. To connect it to finite dimension $\phi(s,a)\in\mathbb{R}^d, \mu(s')\in\mathbb{R}^d$, we use the Monte-Carlo method to approximate it with finite samples,
\begin{align}\label{eq:mc}
    P(s'|s,a)\approx\frac{1}{M}\sum_{i=1}^M \phi_{\omega_i}(s,a) \mu_{\omega_i}(s')q(s').
\end{align}

Introduce vectors $\phi(s,a)$ and $\mu(s')$ such that 
\begin{align}
    \phi(s,a)&:=\frac{1}{\sqrt{M}}\left[\phi_{\omega_1}(s,a), \phi_{\omega_2}(s,a), ..., \phi_{\omega_M}(s,a)\right], \label{eq:phi_rff2linear}\\
    \mu(s')&:=\frac{1}{\sqrt{M}}\left[\mu_{\omega_1}(s'), \mu_{\omega_2}(s'), ..., \mu_{\omega_M}(s')\right].\label{eq:mu_rff2linear}
\end{align}

Then it's straightforward to see that,
\begin{align}\label{eq:appAend}
    \phi(s,a)^\top\mu(s')q(s')&=\frac{1}{M}\sum_{i=1}^M(\phi_{\omega_i}(s,a)^\top \mu_{\omega_i}(s'))\nonumber\\
    &\approx P(s'|s,a).
\end{align}
Hence, Eq.~\ref{eq:p_ebm} can, in principle, yield the motif representation introduced earlier.

\clearpage
\section{Multi-task gridworld dataset}\label{sec:app_gridworld}
\subsection{Dataset}
To generate the dataset, we follow this procedure:  
1) Use soft value iteration to compute the ground truth Q-function for each task: $Q(s,a,t) = r(s,a,t) + \log \sum_a \exp V(s,t)$; 2) Use the resulting Q-function to define the policy: $\pi(a|s,t) = {\exp(Q(s, a, t))}/{\sum_{a'} \exp(Q(s, a', t))}$ and sample trajectories accordingly.

\begin{figure*}[ht]
    \centering
    \includegraphics[width=\textwidth]{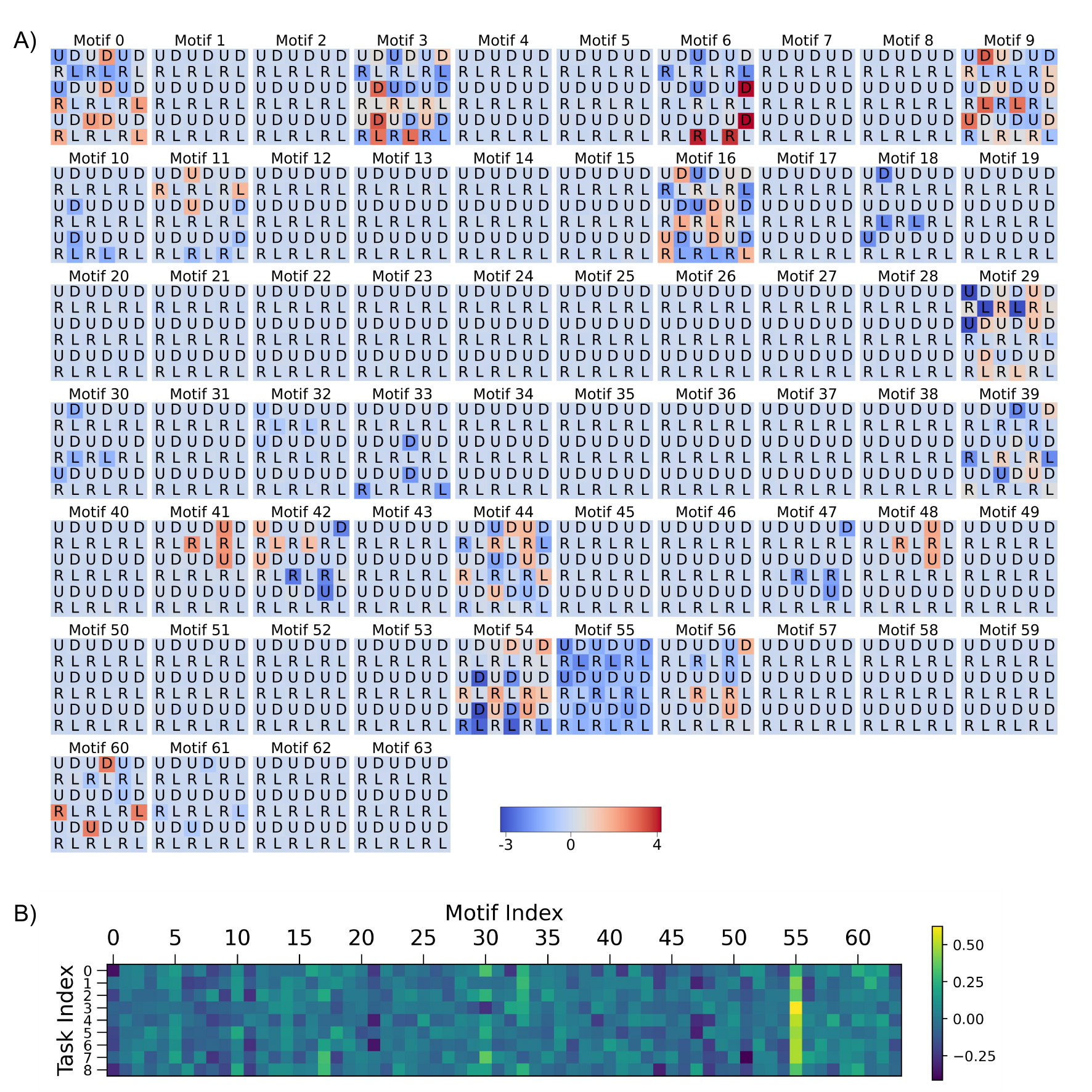}
  \caption{\textbf{A.} State-action maps for all 64 motifs. \textbf{B.} Reward weight $w$ for all 64 motifs.}
  \label{fig:app_gridworld}
\end{figure*}

\subsection{Learned motifs}
We visualize all original 64 motifs (Fig.~\ref{fig:app_gridworld}) introduced in Sec.~\ref{sec:exp1}. It shows some meaningful patterns as mentioned before. For example, motif 0 assigns high values to those $(s,a)$ pairs leading to the middle-middle grid and the bottom-middle grid, and assigns low values to the up-left grid and the up-right grid. Thus, it is employed negatively in Task 0 (up-left reward) and Task 2 (up-right reward). However, Task 4 (middle-middle) didn't use this motif and used motif 39 negatively instead. The complex many-to-many relationship between motifs and tasks informs us of the redundancy in the original motifs, which inspires us to use PCA to analyze the principal components of the motif space and simplify the motif weights. It could be seen from the comparison between Fig.~\ref{fig:exp1}D and Fig.~\ref{fig:app_gridworld}A that principal components are a less redundant description of the motif space.

\clearpage
\section{Animal navigation behavior dataset}\label{sec:app_labyrinth}


\subsection{Learned motifs}
\begin{figure*}[t]
    \centering
    \includegraphics[width=1\textwidth]{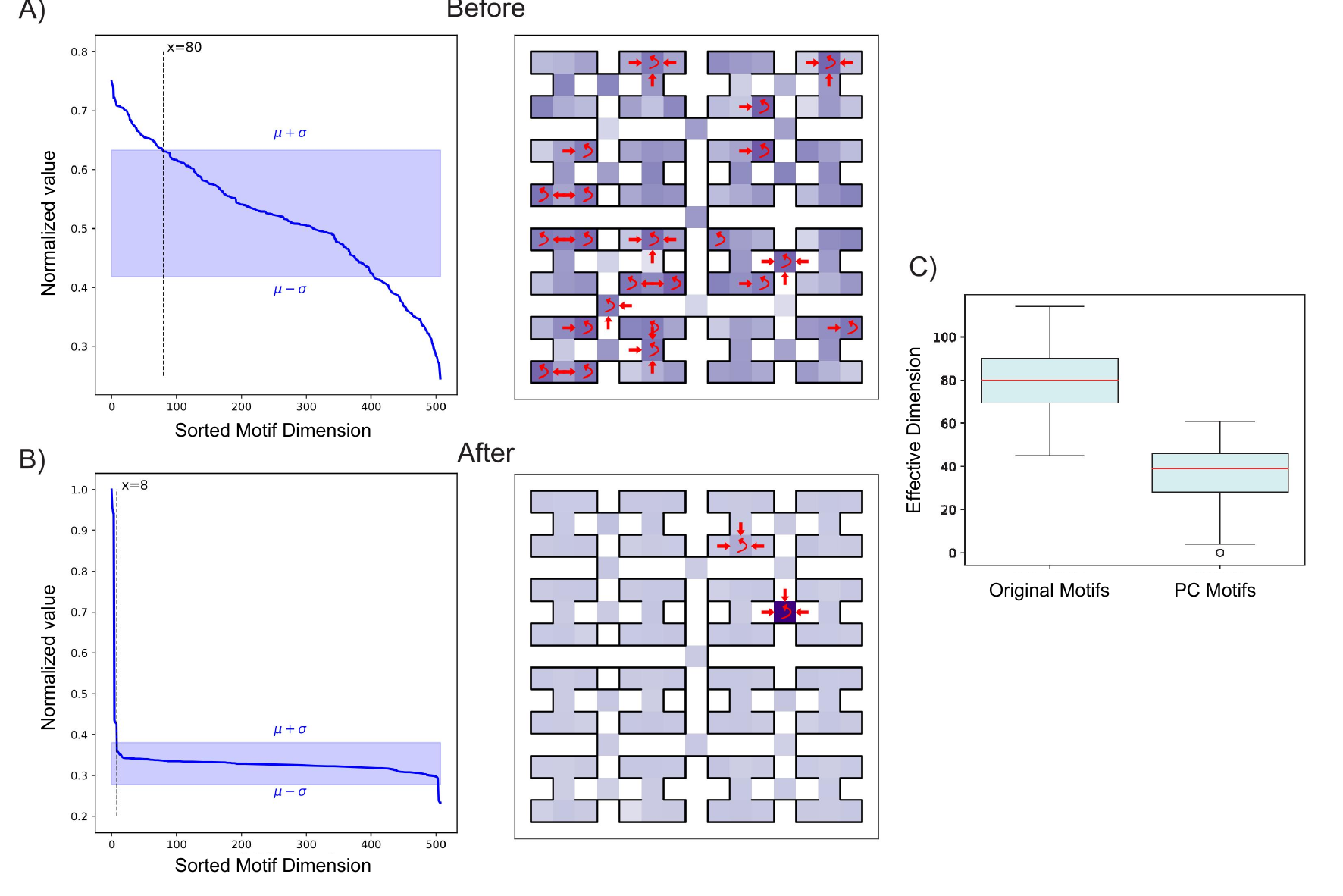}
  \caption{\textbf{A.} Left. The value for each $(s,a)$ pair in motif 0 before PCA. Right. The most important $(s,a)$ pairs. \textbf{B.} Left. The value for each $(s,a)$ pair in motif 0 after PCA. Right. The most important $(s,a)$ pairs. \textbf{C.} Boxplot for the effective dimensions before and after PCA. }
  \label{fig:app_labyrinth}
\end{figure*}

In the original motifs of the labyrinth environment, multiple $(s,a)$ pairs are simultaneously activated, so it is rather hard to analyze which $(s,a)$ pairs are the most important ones that could represent the focus and function of the motif. Given the redundancy of the motif sets, as in Appendix~\ref{sec:app_gridworld}, we perform PCA to analyze the principle components of the motif space and simplify the motif representation. To show the effect of PCA, we plot one motif (motif 0) before (Fig.~\ref{fig:app_labyrinth}A right) and after PCA (Fig.~\ref{fig:app_labyrinth}B right). Basically, we only want to show the most important pairs in one map and do not want low-value pairs to disturb the visualization. To determine how many $(s,a)$ pairs are important, we sort the $(s,a)$ pairs based on the value $\phi(s,a)$ in motif 0, i.e., the first feature of the output of $\phi(s,a)$ (Fig.~\ref{fig:app_labyrinth}A left and B left). It could be seen straightforwardly that after PCA, the motif becomes more concentrated on several $(s,a)$ pairs. We calculate the mean $\mu$ and standard deviation $\sigma$ across all motifs and all dimensions and take $\mu +\sigma$ as the threshold, above which $(s,a)$ pairs are deemed the most important ones and are shown on the right. We show 80 pairs before PCA and 8 pairs after PCA. The number of the most important pairs in each motif is called the ``effective dimension.'' The effective dimension is calculated across all motifs (Fig.~\ref{fig:app_labyrinth}C). Paired t-test ($p=1.3\times 10^{-51}$) shows that there exists a significant decrease of effective dimensions after PCA. So the map becomes more distinct and functionally separated.

\clearpage
\section{Animal free-moving behavior dataset}\label{sec:app_keymoseq}
\subsection{Dataset and training}

We split the full dataset into training and test trajectories in an 8:2 ratio. We first learn both $f$ and $u(t)$ on the training set. Then, given the learned $f$, we estimate $u(t)$ on the test set. Here, $f$ is a time-invariant model parameter shared across all time, while $u(t)$ is a time-dependent variable that must be inferred separately for each test trajectory and cannot be transferred from training.

\subsection{Learned motifs}
We show all motifs learned from the 200-timestep video clip of the free-moving mouse mentioned in Sec.~\ref{sec:exp3}. We have completed the visualization of those motifs that were previously omitted due to their perceived lack of importance. Due to the increased number of displayed motifs, we had to renumber each motif. We show the present motif number above the motif motion field figure, and previous numbers (if applicable) in the parentheses following the present number.

\begin{figure*}[ht]
    \centering
    \includegraphics[width=1\textwidth]{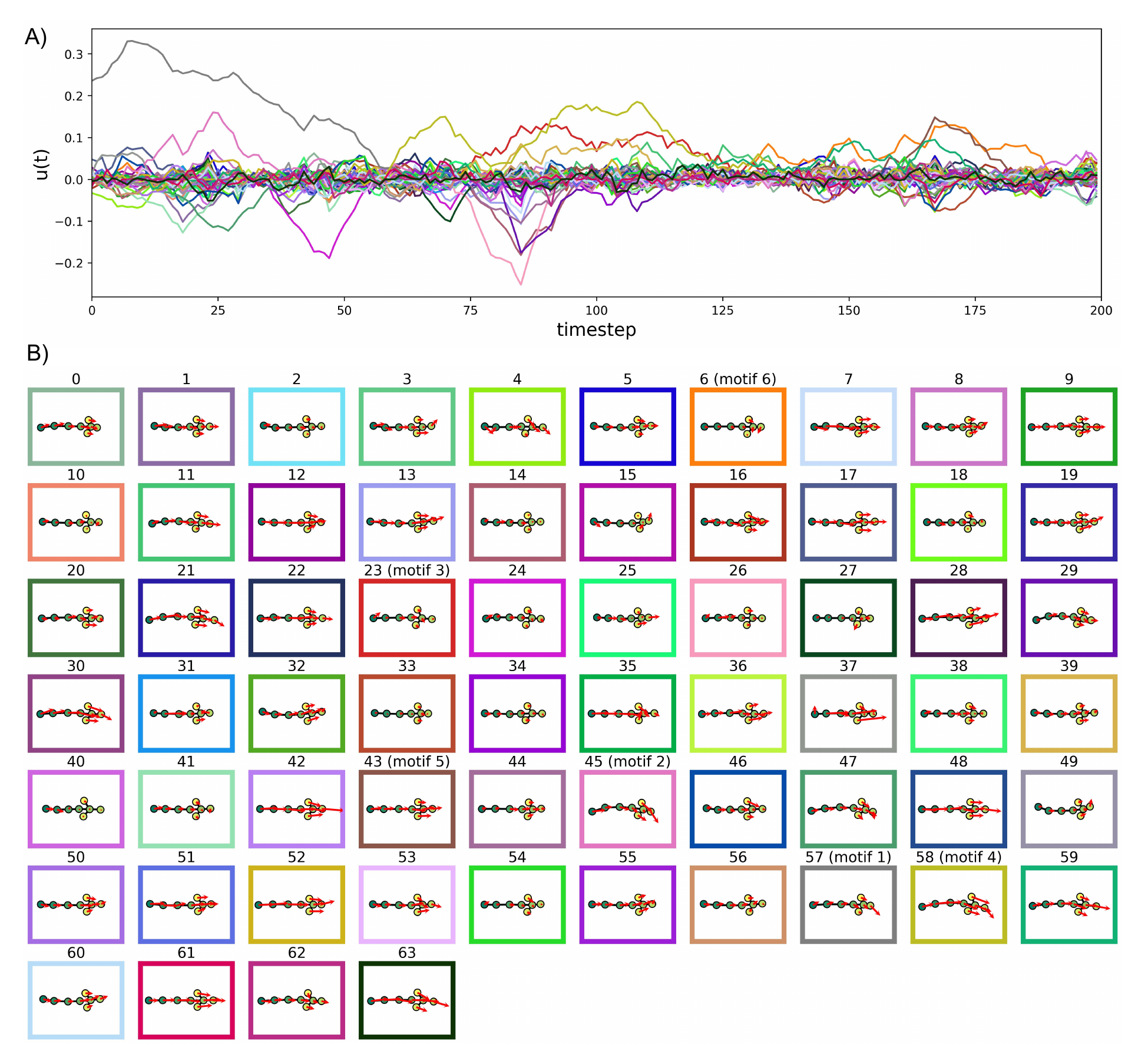}
  \caption{\textbf{A.} Policy weight $u(t)$ for all 64 motifs. \textbf{B.} The motion field for all 64 motifs learned from the video, computed by averaging the states and actions that most strongly activate each motif. }
  \label{fig:app_keymoseq}
\end{figure*}

\clearpage
\section{Macaque drawing dataset}
\label{sec:app_macaque_drawing}

\paragraph{Task and motivation.}
We evaluate MCD on the macaque drawing dataset of
Tian et al.~\cite{tian2026neural}, in which two
macaques, Diego and Pancho, copied geometric figures
on a touchscreen. The original study investigated
whether familiar drawing strokes, or
\emph{primitives}, were reused when the animals
produced novel characters.

We ask two related questions: whether the learned
MCD representation separates familiar primitive
strokes into recognizable groups, and whether these
groups can be identified within unseen characters.
We address the first question by clustering isolated
training strokes with K-means, and the second by matching
novel-character strokes to the resulting templates.
Our analysis uses Diego's behavioral trajectories. No neural recordings enter MCD training or evaluation.

\paragraph{Training, calibration, and test data.}
We construct four non-overlapping subsets from
Diego's released trajectories
(Table~\ref{tab:macaque-subsets}).

\begin{table}[t]
\centering
\caption{Data subsets constructed from Diego's
drawing trajectories. Calibration and test strokes
are excluded from MCD training.}
\label{tab:macaque-subsets}
\small
\begin{tabular}{@{}p{0.25\linewidth}p{0.39\linewidth}p{0.29\linewidth}@{}}
\toprule
Subset & Contents & Purpose \\
\midrule
Isolated primitive basis
& 413 isolated strokes from 19 predefined single-shape types
& MCD training and K-means template construction \\

Practised characters
& 9,981 strokes from 1,071 character identities
& MCD training \\

Novel characters
& 2,040 strokes from 665 character identities
& Held-out evaluation \\

Primitive controls
& 3,577 strokes from separate isolated and grid trials
& Similarity-threshold calibration \\
\bottomrule
\end{tabular}
\end{table}

The first two subsets are used to train MCD.
After training, the isolated primitive basis is
used to construct templates by K-means clustering.
The primitive controls contain strokes from separate
single-primitive and grid-arranged primitive trials.
They are reserved for calibrating match thresholds,
not for fitting MCD or the templates. The novel-character
strokes are used only for evaluation. Thus, the
representation, clusters, and thresholds are all
established without using the novel-character test set.

\paragraph{Preprocessing and task definition.}
We downsample each stroke to 25~Hz by retaining
every twentieth sample of the original 500~Hz
trajectory. We express its two-dimensional
coordinates relative to its starting position
and apply a coordinate scale estimated from
training trajectories only.

For transition $t$ of stroke $r$, the state
$s_{r,t}$ is its stroke-local position and the
action is the displacement to the next sampled
position, $a_{r,t}=s_{r,t+1}-s_{r,t}$. Stroke boundaries are
preserved, and transitions never connect strokes.

MCD uses task-conditioned policy coefficients.
For practised-character trials, task identity is
the character identity, shared by all strokes of
a given character. The isolated-stroke trials
contribute 19 additional single-shape task
identities, yielding $1{,}071+19=1{,}090$ training
tasks. Task $k$ conditions the policy coefficients
$u^{(k)}$ in
$Q(s,a\mid k)=\phi(s,a)^{\mathsf T}u^{(k)}$,
while $\phi(s,a)$ is shared across tasks. The primitive label from the dataset is removed from training.

\paragraph{MCD training.}
We train a shared 64-dimensional motif
representation $\phi(s,a)\in\mathbb{R}^{64}$
with task-conditioned policy learning.
For each update, we sample a training task
uniformly, then a stroke within the task, and
finally a transition within the stroke. This
reduces domination by long strokes or heavily
represented tasks. We use the checkpoint after
1,000,000 training steps, with batch size 128,
feature learning rate $10^{-4}$, policy learning
rate $3\times10^{-4}$, and discount factor $0.9$.
The feature network has no hidden layer.
We freeze the learned representation before
constructing templates or evaluating novel strokes.

\paragraph{Stroke embeddings and unsupervised templates.}
We represent each stroke $r$ by averaging its
transition-level motif features over its $T_r$
transitions:
\begin{equation}
z_r=\frac{1}{T_r}\sum_{t=1}^{T_r}
\phi(s_{r,t},a_{r,t}).
\label{eq:app-macaque-stroke}
\end{equation}

We apply K-means to the 413 isolated training
stroke embeddings with $K=19$, matching the size
of Diego's predefined primitive reference library.
For each discovered cluster $c$, the reference
template is the mean of its member embeddings:
\begin{equation}
\mu_c=\frac{1}{|\mathcal{B}_c|}
\sum_{r\in\mathcal{B}_c}z_r,
\label{eq:app-macaque-centroid}
\end{equation}
where $\mathcal{B}_c$ contains the isolated
strokes assigned to cluster $c$.

Clustering uses the mean-pooled embeddings
without additional normalization or whitening.
We run ten predetermined clustering seeds,
with 50 K-means initializations per seed.

\paragraph{Agreement with predefined primitive types.}
After clustering, to assess whether the discovered clusters
correspond to recognizable primitives, we compare
them with the predefined types of
the isolated strokes. We use Hungarian matching
to obtain a one-to-one alignment between cluster
identities and primitive types.

Averaged over the ten clustering seeds, MCD
achieves $89.8\%$ Hungarian-aligned agreement,
an adjusted Rand index (ARI) of $0.840$,
and normalized mutual information (NMI) of
$0.917$. 

These scores are evaluated on the same 413
isolated strokes used to fit K-means. They
measure agreement of the discovered partition
with the reference labels. ARI is adjusted for chance and has
an expected value of zero under random independent
partitions.

\paragraph{Measuring stroke similarity.}
We compare each stroke embedding with all
19 templates using cosine-based similarity:
\begin{equation}
S(z,\mu)=
\left(
2-\frac{z^\mathsf{T}\mu}
{\|z\|_2\|\mu\|_2}
\right)^{-1}.
\label{eq:app-macaque-similarity}
\end{equation}
Equivalently, $S(z,\mu)=1/(1+d_{\cos}(z,\mu))$,
where $d_{\cos}$ is cosine distance.
Higher similarity indicates a closer match.

Each stroke is assigned to its nearest template:
\begin{equation}
\hat c_r=
\operatorname*{arg\,max}_{c\in\{1,\ldots,19\}}
S(z_r,\mu_c).
\label{eq:app-macaque-assignment}
\end{equation}

Every stroke has a nearest template, even
when that template is a poor match.
We therefore additionally require its similarity
to pass a template-specific threshold.

\paragraph{Calibrating high-quality match thresholds.}
Following the original study's nearest-template
matching logic, we estimate a separate similarity
threshold $\tau_c$ for each template.
The thresholds use only the 3,577 independent
primitive-control strokes, not MCD training
strokes or novel-test strokes.

We first embed each control stroke and assign
it to its nearest template. For template $c$,
we collect the similarities of the controls
assigned to it:
\begin{equation}
\mathcal{S}_c=
\left\{
S(z_r,\mu_c):
r\in\mathcal{C},\ \hat c_r=c
\right\},
\label{eq:app-macaque-controls}
\end{equation}
where $\mathcal{C}$ is the calibration set.

We remove similarities below
$Q_{1,c}-3.5\,\mathrm{IQR}_c$,
where $Q_{1,c}$ is the first quartile and
$\mathrm{IQR}_c$ is the interquartile range
of $\mathcal{S}_c$.
The cutoff is then the 2.5th percentile of
the remaining similarities:
\begin{equation}
\tau_c=
\operatorname{Quantile}_{0.025}
\left(\mathcal{S}_c^{\mathrm{filtered}}\right).
\label{eq:app-macaque-threshold}
\end{equation}

This procedure estimates the similarity range
observed for independent primitive controls,
while preventing extreme low-similarity outliers
from making the cutoff overly permissive.
All thresholds are fixed before test evaluation.

\paragraph{High-quality match classification.}
A novel-character stroke is accepted as a
high-quality (HQ) match if its similarity
to its nearest template reaches that
template's cutoff:
\begin{equation}
h_r=
\mathbf{1}\left[
S(z_r,\mu_{\hat c_r})
\geq\tau_{\hat c_r}
\right].
\label{eq:app-macaque-hq}
\end{equation}

Thus, HQ indicates that a stroke passes the
calibrated similarity criterion for its assigned
template. 

\paragraph{Comparison methods.}
We compare trained MCD with two references.

\emph{Untrained MCD} uses the same encoder
architecture with frozen random parameters.
We apply the same K-means template construction,
cosine-based matching, and independent threshold
calibration as for trained MCD.
Both variants use ten clustering seeds.

\emph{Label-based DTW} uses the original study's
released primitive templates and velocity-based
dynamic time warping (DTW) procedure.
Unlike MCD's K-means templates, this reference
library uses predefined primitive labels.

For every method, we independently calibrate
its thresholds using the same 3,577 controls
and evaluate the same novel-test strokes.
This matches the evaluation dataset and
the nearest-template acceptance rule, while
allowing each method to use its own templates,
similarity function, and calibration cutoffs.
The label-based DTW baseline uses predefined
labels, and is compared with MCD which discovers primitive in an unsupervised way.

\paragraph{Full novel-character test set.}
The full test set contains 2,040 strokes
from 665 novel character identities drawn
by Diego. Its HQ rate is the fraction of
test strokes accepted:
\begin{equation}
\mathrm{HQ}_{\mathrm{full}}=
\frac{1}{|\mathcal{T}|}
\sum_{r\in\mathcal{T}}h_r,
\label{eq:app-macaque-hq-full}
\end{equation}
where $\mathcal{T}$ contains all novel-test
strokes.

\paragraph{Shared-character subset.}
For the main-text figure, we additionally
restrict evaluation to character identities
drawn by both Diego and Pancho.

Using the released experimental metadata,
we first identify groups defined by animal,
recording date, and task set that contain
both novel and practised character conditions.
Among these eligible groups, we find the
exact character identities represented for
both animals. We then retain Diego's
held-out novel-test strokes with those
identities. This gives 394 strokes from
135 character units.

Pancho's data are used only to identify
shared character identities. His trajectories
are not used to train or evaluate the Diego
MCD model.

For this subset, we calculate the HQ fraction
within each character unit, using the released
character-group identifier, and average these
fractions with equal weight:
\begin{equation}
\mathrm{HQ}_{\mathrm{shared}}=
\frac{1}{|\mathcal{G}|}
\sum_{g\in\mathcal{G}}
\left(
\frac{1}{|\mathcal{T}_g|}
\sum_{r\in\mathcal{T}_g}h_r
\right),
\label{eq:app-macaque-hq-shared}
\end{equation}
where $\mathcal{G}$ contains the selected
character units and $\mathcal{T}_g$ contains
the test strokes in unit $g$.

This gives equal weight to each character
unit rather than to each stroke.
The full and shared results therefore differ
in both the selected drawings and the averaging
rule. Templates and calibrated thresholds
remain unchanged between subsets.

\paragraph{Aggregation and uncertainty.}
For MCD and untrained MCD, point estimates
are averaged over the ten clustering seeds.
For the full test set, we first average each
stroke's HQ indicator across seeds and then
calculate the overall stroke-weighted rate.
For the shared subset, we average the
within-character HQ fractions across seeds
and then across character units.

We obtain 95\% bootstrap intervals by resampling
whole character identities for the full test
set and whole character units for the shared
subset, with replacement. All strokes within
a sampled character or unit remain together.
The full-set bootstrap recomputes the
stroke-weighted estimator, whereas the
shared-subset bootstrap recomputes the equally
weighted mean of character-unit HQ fractions.

The bootstrap reflects sampling variation
across evaluation characters, conditional on
the trained representation and the templates
and calibration thresholds obtained for each
clustering seed. It does not quantify uncertainty
from independently retraining MCD. The plotted
intervals must therefore not be interpreted as
variation across MCD training runs.

\paragraph{Results and interpretation.}
Table~\ref{tab:macaque-hq-results} reports the
HQ rates and 95\% bootstrap intervals.

\begin{table}[t]
\centering
\caption{Primitive-matching HQ rates on
Diego's novel drawings, with 95\% bootstrap
intervals. All methods use the same evaluation
strokes within each subset. MCD and untrained
MCD use K-means templates without primitive-type
labels. Their results are averaged over ten
clustering seeds.}
\label{tab:macaque-hq-results}
\small
\begin{tabular}{@{}lcc@{}}
\toprule
Method & Full test set & Shared subset \\
\midrule
Label-based DTW
& 85.0\% (83.5--86.6\%)
& 90.3\% (87.0--93.2\%) \\
MCD
& 80.2\% (78.5--81.8\%)
& 86.9\% (83.6--90.1\%) \\
Untrained MCD
& 65.1\% (63.1--67.1\%)
& 70.9\% (66.2--75.5\%) \\
\bottomrule
\end{tabular}
\end{table}

On the shared-character subset, MCD achieves
an HQ rate of $86.9\%$, compared with $90.3\%$
for label-based DTW and $70.9\%$ for untrained
MCD. The same ordering holds on the full test
set, with rates of $85.0\%$, $80.2\%$, and
$65.1\%$ for DTW, MCD, and untrained MCD,
respectively.

The two evaluations provide complementary
evidence: K-means groups isolated-stroke
embeddings into clusters that largely agree
with the original primitive types, and the
same clusters cover a high fraction of strokes
within novel character compositions. The
trained representation also exceeds the
random-encoder baseline under the same
clustering and calibration procedure. These findings support the presence and
transfer of reusable stroke structure in
MCD's learned representation.

\clearpage
\section{External validity}\label{sec:app_external_validity}

MCD's result of the navigation labyrinth dataset is based on the task segmentation of \cite{ke2025inverse}. There remains two questions: (1) how do other models perform on the same segmentation; and (2) whether the ground truth from \cite{ke2025inverse} is valid. To further validate our results, we evaluated two additional inverse reinforcement learning baselines: (1) IQ-Learn (IQL, \cite{garg2021iq}), where we use the same segmentation from \cite{ke2025inverse} to infer a reward map for each task; and (2) Dynamic-IRL (DIRL, \cite{ashwood2022dynamic}), where without relying on this segmentation, we infer a time-varying mixture of reward functions directly from their curated data. We use a tabular version of IQL here just like MCD. For comparison, we constructed two auxiliary reward maps: a perfect reward map with artificially placed rewards, and a random baseline map whose entries are sampled from a standard normal distribution. In the perfect map, the water-seeking task reward function has a peak of 1 at the water port. The home-seeking task reward function has a peak of 1 at the home location. And the exploration task reward function has a uniform low reward of 0.3 at every location.

\begin{figure}[ht]
    \centering
    \includegraphics[width=0.8\textwidth]{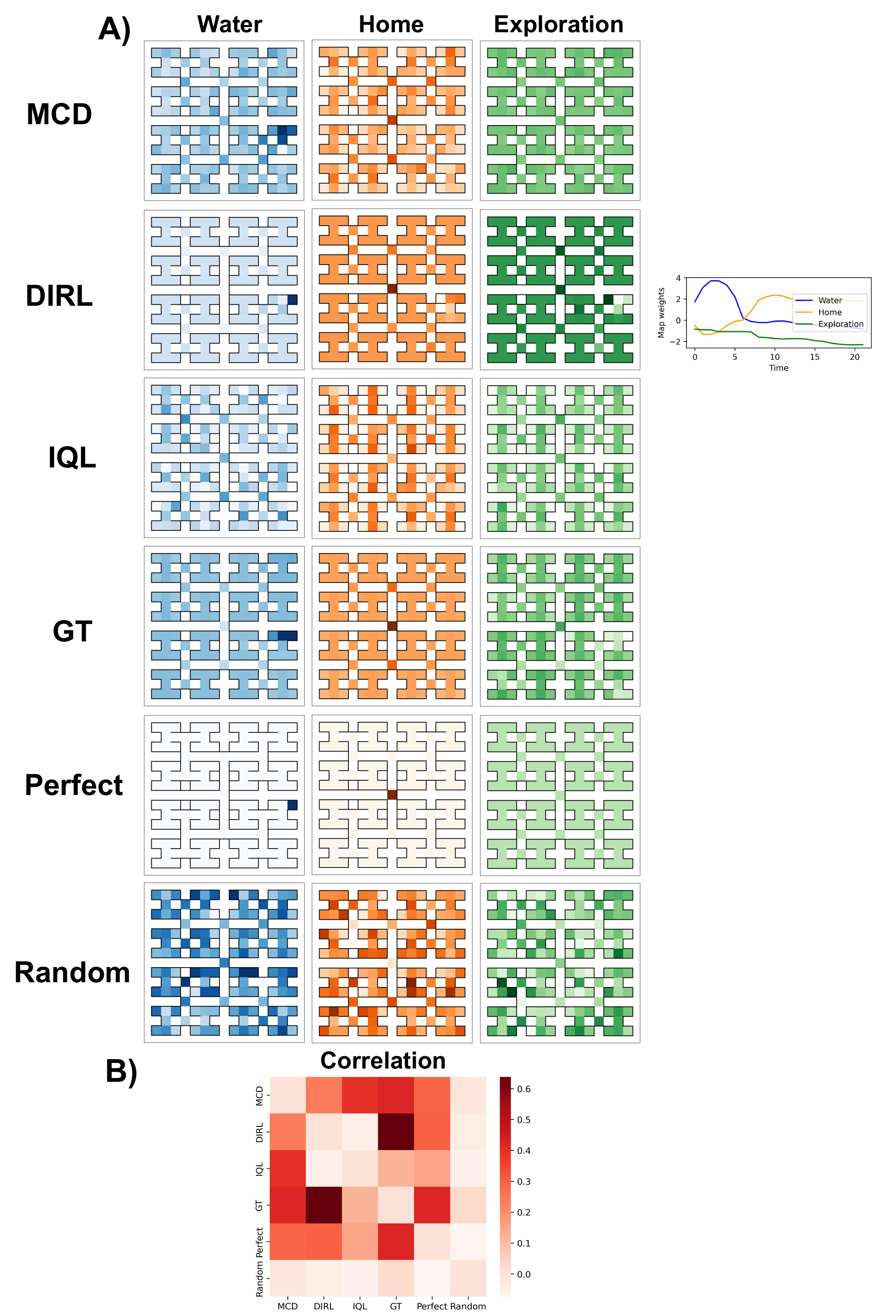}
  \caption{\textbf{A.} Reward maps inferred by different models. The maps from DIRL are ranked based on the time-varying weights (right at the corresponding row). "Water" weight increases and then decreases; "home" weight remains low and then increases; and "exploration" weight remains low. \textbf{B.} Correlations across the reward maps.}
  \label{fig:app_keymoseq_external}
\end{figure}

Fig.~\ref{fig:app_keymoseq_external}A shows all inferred maps, and Fig.~\ref{fig:app_keymoseq_external}B reports the Pearson correlation coefficients across all maps. There are three important conclusions. (1) The strong alignment between the MCD-recovered maps, the ground-truth maps from \cite{ke2025inverse} (GT), and the perfect manual map demonstrates that MCD successfully recovers the underlying motivational structure. (2) MCD achieves consistently higher correlation with both GT and the perfect map than IQL under the same task segmentation, suggesting that MCD captures the animal’s latent motivation more accurately than the IQL baseline. (3) The high correlation between GT, the perfect map, and DIRL provides external validation for the quality of the GT annotations themselves. 

Note that although DIRL achieves higher correlations with the ground-truth maps, it is not suitable for our setting. DIRL requires stereotyped, repeated sequences of \emph{identical} length and therefore depends on carefully curated datasets in which such fixed-length segments can be extracted (not the same dataset as we use, but the same environment). In contrast, our datasets—especially the continuous animal free-moving dataset—exhibit substantial heterogeneity and do not contain repeated trajectories of uniform duration. We are also concerned that such curation would introduce bias and distort the natural statistics of the behavior. Furthermore, DIRL relies on \emph{tabular} value iteration to backpropagate gradients from the policy to the reward, which is fundamentally incompatible with the continuous state and action spaces present in the animal free-moving behavior data.

\clearpage
\section{Lunar Lander dataset}

We test MCD on a synthetic continuous Lunar Lander environment. In this environment, the agent controls a rocket descending toward the lunar surface using left, right, and main engines. Each action fires one of the engines, consumes fuel and affects the lander's pose. The final stable landing requires the agent to balance energy efficiency and stability.

\paragraph{State variables.}
At each time step, the environment state is represented by an 8-dimensional vector
\[
(x, y, v_x, v_y, \theta, \dot{\theta}, \text{leg}_L, \text{leg}_R),
\]
where $(x,y)$ denote position, $(v_x,v_y)$ linear velocities, $\theta$ the lander’s angle, $\dot{\theta}$ its angular velocity, and $\text{leg}_L,\text{leg}_R \in \{0,1\}$ indicate ground contact of the left and right landing legs. The \emph{contact} is defined as whether either leg is in contact with the ground. 

The agent selects a discrete action $a_t \in \{0,1,2,3\}$ at each time step, corresponding to:
\begin{itemize}
    \item $0$: do nothing,
    \item $1$: fire the left orientation engine,
    \item $2$: fire the main engine,
    \item $3$: fire the right orientation engine.
\end{itemize}

Rewards increase when the lander achieves a smooth and upright touchdown with minimal oscillation, and decrease when tilting, excessive acceleration, or rolling happens.

\begin{figure}[ht]
    \centering
    \includegraphics[width=0.7\textwidth]{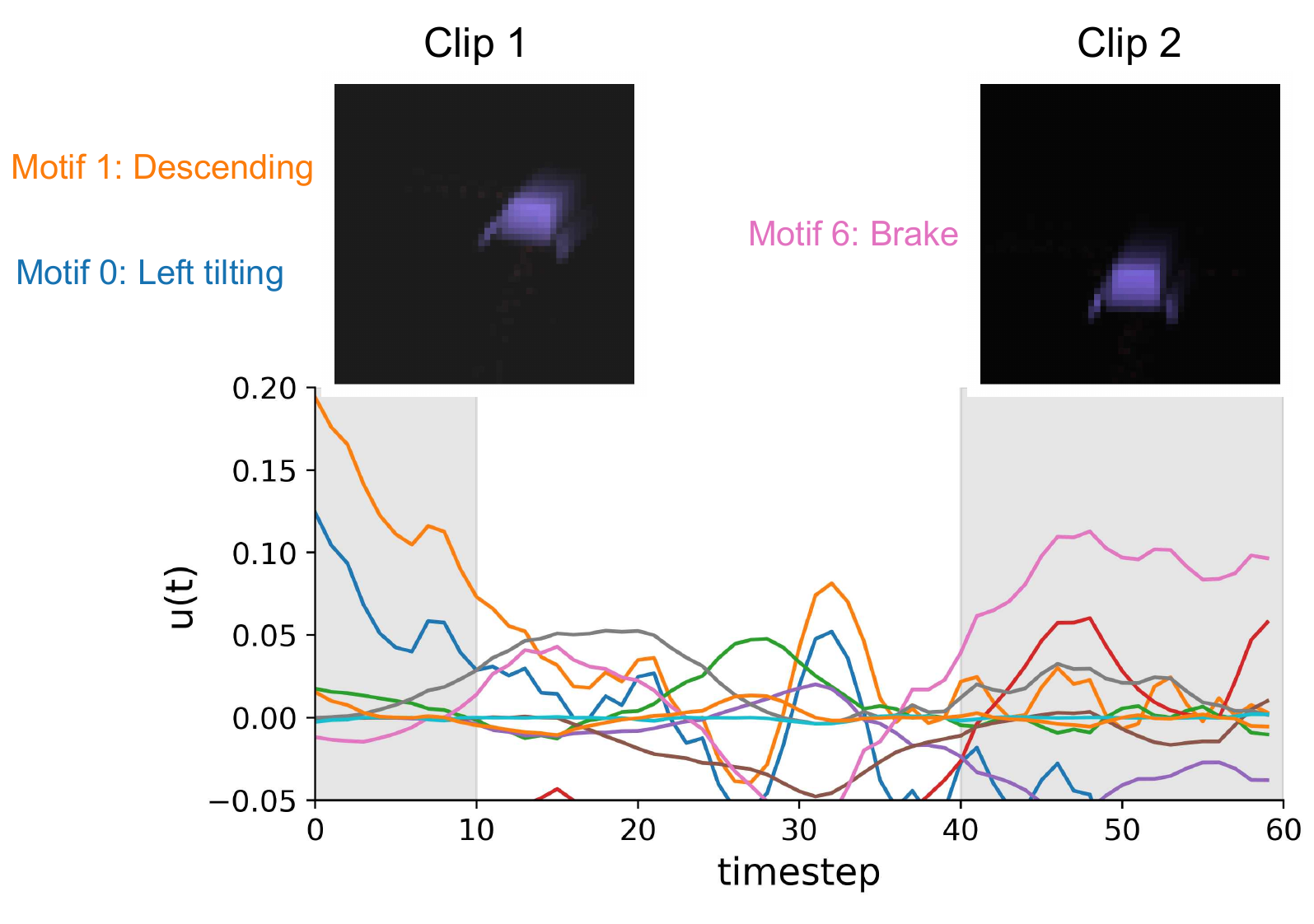}
  \caption{Motif weights for a clip from the lunar lander dataset. }
  \label{fig:app_lunarlander_u}
\end{figure}

We use the continuous formulation in Sec.~\ref{sec:continuous_version} to learn motif representations $\phi(s,a)$ and time-varying weights $u(t)$ (Sec.~\ref{sec:understanding_behavior}). This dataset contain trajectories generated by a trained agent ($N=62544$). To encourage temporal smoothness, we place a Gaussian random-walk prior on $u(t)$: $u(t)\sim\mathcal{N}(u(t-1),\sigma^2 I).$ We optimize $u(t)$ after learning motifs. The motif dimension is set to $d=64$.

We select a representative clip to illustrate interpretability. The first segment exhibits rapid descent with a left tilt, while the second shows braking with smoother descent. This transition is captured by a mixture of two motifs in the first segment and the dominance of Motif 6 in the second (Fig.~\ref{fig:app_lunarlander_u}). We attach the video in the supplementary material. 

\begin{figure}[ht]
    \centering
    \includegraphics[width=\textwidth]{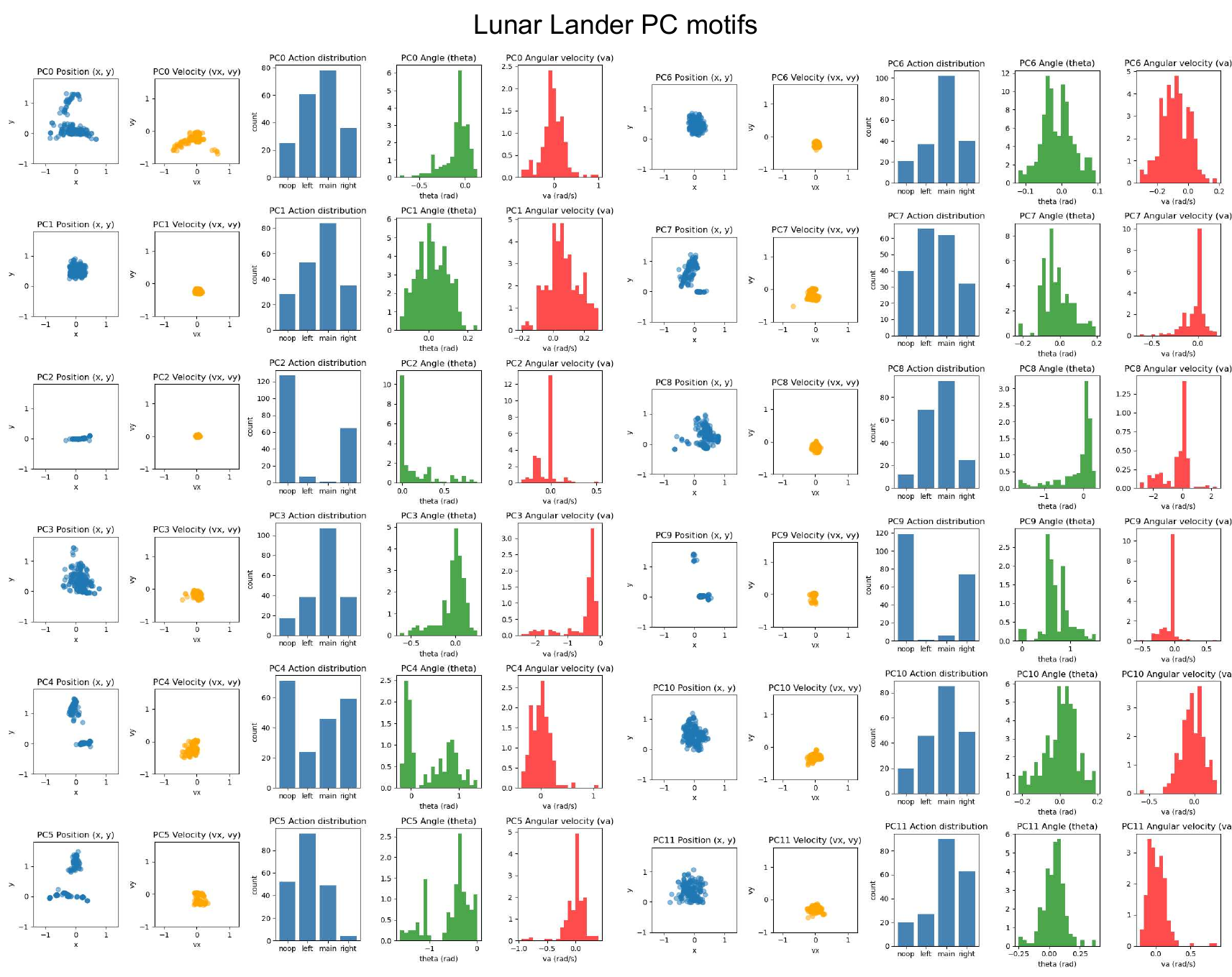}
  \caption{For each PC motif, we collect the data entries that most strongly activate it, and show the distribution of positions, velocities, angle, and angle velocities. }
  \label{fig:app_lunarlander_motif}
\end{figure}

To interpret the motifs, we collect state–action pairs that maximally activate each motif, carry out PCA, and visualize the PC motifs' distributions in Fig.~\ref{fig:app_lunarlander_motif}.

\label{appendix:lunarlander}

\clearpage
\section{Evaluation details on a human-annotated dataset.}\label{sec:appendix_human_label}

We verify our method on a supervised behavior label dataset. Due to the substantial cost and labor demands of manual annotation, large-scale, high-quality animal behavior datasets with human labels remain limited. Commercial annotation tools (e.g., HomeCageScan) are also expensive, further constraining availability. To support quantitative evaluation, we therefore included a human-annotated subset from the freely moving animal behavior dataset we mentioned last section. Specifically, we selected 200 short video clips (10 recording sessions; in each session we randomly sampled 20 clips, with 200 frames each). Each clip was manually annotated with behavior segments for each frame, resulting in 200 per-frame labels across six categories: \texttt{Walking}, \texttt{Sniffing/Grooming}, \texttt{Turning left}, \texttt{Turning right}, \texttt{Rearing}, and \texttt{Resting}. Manual annotation required approximately 40 hours of expert effort. The dataset is split into a training set and a test set with a ratio $8:2$.

As before, we evaluated four models on this dataset: MCD, Keypoint-MoSeq, SemiSeg, and OPAL. Because the latent variables or syllables inferred by these models do not necessarily align with human-defined labels, we trained an additional decoder to map model-specific latent representations to the ground-truth categories. For each model, the decoder input was the motif weights $u(t)$ (MCD), the inferred one-hot syllable (Keypoint-MoSeq), or the latent embedding $z$ (SemiSeg and OPAL). The decoder was implemented as a two-layer neural network with ReLU activation, optimized using the Adam optimizer and cross-entropy loss (learning rate $=0.001$, hidden size$=100$), following standard settings in the scientific computation package \texttt{scikit-learn}. The resulting classification accuracies on the held-out test set (Fig.~\ref{fig:exp3}A) demonstrate that MCD substantially outperforms all baselines, achieving near-perfect accuracy. This indicates that the motifs recovered by MCD align better with actual animal behavior dynamics.

\clearpage
\section{Connections to relevant algorithms in RL}\label{appendix:rl_related_work}

\textbf{Motor primitives learning.} Animal behavior research has focused on identifying motor primitives directly from large-scale recordings. Supervised classification~\citep{marks2022deep,segalin2021mouse}, clustering-based analysis~\citep{hsu2021b,berman2014mapping}, and HMM-based methods~\citep{wiltschko2015mapping,weinreb2024keypoint,luxem2022identifying, whiteway2021semi} label behavior into discrete syllables and analyze neural correlates of transitions. While these approaches reveal stereotyped motifs such as walking or grooming, they typically treat primitives as rigid and exclusive, overlooking the continuous, compositional structure of behavior and its adaptation across contexts. 

In robotics, movement primitives have long been studied~\citep{paraschos2013probabilistic, saveriano2023dynamic, lioutikov2017learning}. ProMP~\citep{paraschos2013probabilistic} extracts primitives from demonstrations but requires labeled skills/motifs, while ProbS~\citep{lioutikov2017learning} jointly infers behavioral segmentation and a primitive library via Expectation-Minimization (EM), enabling unsupervised skill discovery. These approaches resemble dynamics-based behavior segmentation in animal behavior research. 

All methods mentioned here lack an RL perspective that links primitives to policies and reward-driven behavior. Our framework instead leverages offline RL-based imitation learning to uncover motif-based policies that capture both modular primitives and their sequential composition, and provides a principled way to explain why behaviors occur, not just how they unfold, and how reusable motifs/skills contribute to them in a generative decision-making process.

\textbf{Offline Imitation learning.} Offline imitation learning presents the problem of learning a policy from fixed demonstrations when access to environments is impossible. Simple behavior cloning can be performed offline, but fails to generalize well in some cases because it does not consider dynamics or environment structure limits. ValueDICE~\citep{kostrikov2019imitation} considers the dynamics of the training data and learns a policy that minimizes the KL-divergence between the state-action occupancies generated by the policy and of the original dataset. But the adversarial optimization of policy and Q-functions introduces instability in training. IQL~\citep{garg2021iq} avoids adversarial training by directly parameterizing the policy in terms of the $Q$-function, $\pi(a|s)= \exp(Q(s,a))/Z$. However, all of these algorithms learn policies as unstructured functions. As the animal behavior structure is highly modular and stereotyped, it is more appropriate to employ a hierarchical motif-based policy to model the data.

\textbf{Motif/Skill discovery.} Unsupervised skill discovery (DIAYN~\citep{eysenbach2018diversity}, BeCL~\citep{yang2023behavior}, DADS~\citep{sharma2020dynamics}, InfoGAIL~\citep{li2017infogail}, Directed-InfoGAIL~\citep{sharma2018directed}, SkillBlender~\citep{kuang2025skillblender}, ASE~\citep{peng2022ase}) to find a high-level abstraction of actions has been an effective strategy for online RL and imitation learning. However, they are limited to online settings when the skills can only be refined through interacting with the environment, which restricts their application to large-scale offline datasets. Recent offline skill discovery methods include OPAL~\citep{ajay2020opal}, SPiRL~\citep{pertsch2021accelerating}, and SkiLD~\citep{pertsch2021guided} which use an autoencoder to encode trajectories into latent skills $z$; and PARROT~\citep{singh2020parrot} which uses a flow-based model to learn a behavior prior. In these models, the latent $z$ is later used to generate the policy $\pi(a|s,z)$ in a non-linear way. Our paper, instead, employs a generalized-linear structure of the policy $\pi(a|s,z)\propto\exp(\phi(s,a)^\top u(z=t))$ which provides better interpretability than the policy network. This interpretability is essential for addressing downstream neuroscience questions.

\textbf{Linear structure of environment/policy.}
Another line of work (HILP~\citep{park2024foundation}, FB~\citep{touati2021learning}, USFA~\citep{borsa2018universal}, RaMP~\citep{chen2023self}) based on successor features (SF) also uses a generalized linear structure to model the motif-based policy as $\pi(a|s,z)\propto\exp(F(s,a,z)^\top z)$ where $F(s,a,z)$ is SF under a certain motif $z$. The main concern is that they cannot separate motif representation $z$ from state-action representation in SF, while our work can (Eq.~\ref{eq:softpolicy}). Therefore, their motifs depend on the specific task or timepoint, while our motifs/skills are general representations shared across tasks and timepoints. The idea that task-agnostic motifs are combined adaptively to form new policies aligns better with the need of interpretability and scientific discovery: we would like to look for neural signals responsible for relatively fixed time-agnostic behavior patterns, to help us better understand the animal behavior.

TRAIL~\citep{yang2021trail} adopts a linear decomposable environment as MCD when learning the latent skills. However, the latent-conditioned policy $\pi(a|s, z)$ is not linear, but parametrized by a neural network. SkillBlender~\citep{kuang2025skillblender} adopts a linear decomposable policy, and uses linear combinations of lower-level controller outputs as the final actions $a_t=\sum_i k_ia^i_t$, while we use the linear combinations of different motifs to generate the final state-value functions $Q_t(s,a)=\sum_i \phi_i(s,a)u^i_t$ rather than direct actions. We believe a naive mixture of the low-level controller is less biologically realistic in modeling animal behavior than mixtures of state-value functions. The latter can find supporting evidence in neuroscientific literature \citep{makino2023arithmetic}. Besides, their transition kernel $P(s'|s,a)$ is not linear. 

Compared to them, our work assumes generalized linear structures for both the policy and the environment, providing better interpretability. This shared motif is more fundamental in revealing the basic structure of the animal's intentions.


\textbf{Our contributions.} Most existing skill-learning methods do not apply to the behavior understanding scenario considered here. We study offline data without supervision or task annotations—no explicit goals, labels, or trajectory segmentation—only long, unstructured behavior sequences. The challenge is threefold: (1) to discover the basic motor skills, (2) to determine how these skills compose spatially and temporally within long trajectories in an offline setting, and (3) to ensure interpretability for scientific discovery. These are nontrivial problems beyond the reach of existing skill- or motor-primitive approaches.  MCD addresses these challenges within an offline RL-based imitation learning framework, yielding interpretable motif representations and their continuous compositions.

\clearpage
\section{Identifiability discussion}\label{appendix:identifiability}

In this section we clarify the notion of identifiability and how it applies to our motif-based continuous dynamics (MCD) model. A parameterization is said to be \emph{identifiable} if different parameter values lead to different model predictions. Conversely, if multiple distinct parameterizations yield identical likelihoods or value functions, the model is only identifiable up to an equivalence class (e.g., scaling, rotation, or permutation).

\paragraph{Our model is identifiable at the subspace level.}
In MCD, both the transition operator and the action-value function admit the
spectral form
\[
P(s'|s,a) = \phi(s,a)^\top \mu(s'), \qquad 
Q(s,a)   = \phi(s,a)^\top u.
\]
Although individual components of $\phi$, $\mu$, and $u$ may be rescaled by an invertible linear transformation without altering $P$ or $Q$, the \emph{subspace spanned by the motif functions} is uniquely determined \citep{ren2022spectral}. This is analogous to PCA, where the principal subspace is identifiable even if the basis vectors within that subspace are not uniquely determined. In practice, they converge well.

Moreover, in the continuous version, the identifiability of the new representations $\psi, \nu$ up to orthogonal transformations has been guaranteed by the normalization operation before the output (Appendix~\ref{appendix:hyperparameter}), which also ensures the trainability of the contrastive learning algorithm.  

\paragraph{Comparison to HMMs and latent variable models.}
Classical behavioral segmentation methods (e.g., Keypoint-Moseq as an HMM variant, SemiSeg as an autoencoder variant) are generally not identifiable without strong additional assumptions. Different transition matrices, emission matrices, or latent dynamics parameters can produce indistinguishable observation distributions. These models typically suffer from label-swapping and more severe transform equivalences in the latent space, meaning that even the latent subspace is not uniquely determined. In contrast, our method fixes the ambient representation through the observed state--action pairs and learns a function class whose span is uniquely tied to the transition operator and value function. Thus, while our motif basis is identifiable only up to linear transforms (as is standard for spectral decompositions), the underlying subspace and its behavioral interpretation are stable, reproducible and identifiable.

\paragraph{Summary.} MCD inherits the identifiability properties of spectral RL methods: the motif basis is identifiable up to an invertible transformation, and the motif \emph{subspace}---the structure that determines $P$, $Q$, and $\pi$---is uniquely recoverable. This stands in contrast to latent-variable models such 
as HMMs or SLDS, which in general do not admit identifiable latent subspaces 
without strong, often unrealistic, constraints.

\clearpage
\section{Long-term dependency discussion}\label{appendix:long-term}
Compared with prior approaches, our framework implicitly captures the long-term temporal structure of behavior through the Bellman equation in the reinforcement learning (RL) formulation. The component responsible for encoding long-term and multi-scale dynamics is the motif weight $u(t)$. From the linear decomposition of the state--action value function (Eq.~\ref{eq:q_linear}), 
\[
u(t)= w + \gamma \int V(s') \, \mu(s') \, q(s') \, ds',
\]
it follows that $u(t)$ integrates two sources of information simultaneously: (i) long-horizon structure through the value function $V(s')$, which summarizes discounted future trajectories with an infinite horizon, and (ii) short-horizon structure through the immediate reward parameter $w$. Therefore, this formulation naturally enables modeling of multi-scale temporal dependencies. Empirically, we observe this structure in Fig.4B: some motifs exhibit stable, slowly varying activation profiles (e.g., motif 1, gray curve), whereas others fluctuate rapidly and capture short-term transitions (e.g., motif 2, pink curve). Intuitively, consider a mouse that intermittently sniffs throughout the day but engages in fast running only during a brief morning period. The coefficient $u_i(t)$ for sniffing would rise gradually across long time windows, while the coefficient $u_j(t)$ for running would increase only during short, specific intervals.


\clearpage
\section{Continuous composition versus smoothing of discrete motifs}\label{appendix:motif-continuity}

\paragraph{Question and experimental setup.}
A smoothly changing representation does not necessarily imply continuous
composition. It could instead represent one discrete state at a time, with
smoothing creating mixtures only when the state changes. We test this
explanation using two complementary analyses: where mixing occurs in time,
and how much behavioral information remains after replacing MCD with
discrete states.

We use the human-annotated free-moving mouse dataset, containing $200$ clips
of $200$ frames each at $30$ fps and six behavior categories. We reuse the
trained MCD model and its inferred $64$-dimensional representation $u(t)$. MCD is not retrained for this analysis.

\paragraph{Mixing within stable bouts and at behavior boundaries.}
To measure the mixing of $u(t)$, we normalize their absolute magnitudes into relative component weights:
\begin{equation}
    p_j(t)
    = \frac{|u_j(t)|+\epsilon}
           {\sum_{k=1}^{D}\bigl(|u_k(t)|+\epsilon\bigr)},
    \qquad D=64,\quad \epsilon=10^{-12}.
    \label{eq:continuity-weights}
\end{equation}
and quantify mixing using normalized Shannon entropy and top-1 mass:
\begin{equation}
    H(t) = -\frac{\sum_{j=1}^{D}p_j(t)\log p_j(t)}{\log D},
    \qquad M_1(t)=\max_{j}p_j(t).
    \label{eq:continuity-metrics}
\end{equation}
Entropy is near zero when one component dominates and increases as weight
is spread across components. Top-1 mass measures the share carried by the
largest component.

We call a frame a boundary frame if it lies within $\pm5$ frames
(approximately $\pm167$ ms) of a change in the human behavior label.
All remaining frames are stable-bout interiors. A stable bout means that
the annotated behavior stays the same.

For comparison, we encode the six human behavior labels as one-hot states. We smooth this sequence
within each clip using a Gaussian kernel with $\sigma=3$ frames
($100$ ms). This creates a smoothed discrete labels baseline whose mixing is concentrated
around behavior changes. Its entropy is normalized by $\log 6$, so that
both representations have entropy on a zero-to-one scale.

As shown in Table~\ref{tab:app-continuity-mixing}, the smoothed discrete
baseline is almost entirely dominated by one component inside stable
bouts, with mixing mainly near boundaries. In contrast, MCD has nearly the same entropy in both regions. Its largest component carries approximately $11\%$ of the
absolute weight near boundaries and in stable interiors. This pattern
remains qualitatively unchanged for boundary half-windows of $1$, $3$,
$5$, $10$, and $15$ frames. Thus, MCD's mixing is not concentrated at the
human-annotated behavior transitions.

\begin{table}[htbp]
\centering
\small
\caption{Frame-weighted mixing statistics near behavior changes and within
stable bouts. Boundary frames use a $\pm5$-frame window.}
\label{tab:app-continuity-mixing}
\begin{tabular}{@{}lccc@{}}
\toprule
Representation & Boundary entropy & Interior entropy & Interior top-1 mass \\
\midrule
Continuous MCD $u(t)$ & $0.7915$ & $0.7877$ & $0.1126$ \\
Smoothed discrete labels & $0.2640$ & $0.0062$ & $0.9978$ \\
\bottomrule
\end{tabular}
\end{table}

\paragraph{Does smoothing discrete states preserve behavioral information?}
High entropy alone does not establish that mixing is useful. We therefore
ask whether discrete states with smoothed transitions can preserve the
human-label information available in continuous $u(t)$.

For each decoding split, we fit a $64$-cluster K-means quantizer to the
training-frame representations, without using human labels. Each frame
is then represented by a one-hot code indicating its nearest cluster.
We compare these hard codes with a version smoothed within each clip using
the same $\sigma=3$-frame Gaussian kernel. The number of discrete states
matches the $64$ components of MCD. Unlike the label-based diagnostic above,
these discrete codes are obtained from MCD rather than from the decoding
targets.

We use the same decoding protocol for all three representations: an MLP
with one hidden layer of $100$ units, matched random $80/20$ frame-level
train/test splits, and ten random seeds. Table~\ref{tab:app-continuity-decoding}
reports mean accuracy and standard deviation across seeds. The
majority-class baseline is the accuracy obtained by always predicting the
most frequent behavior label; it is $0.5566$ for these data.

\begin{table}[htbp]
\centering
\small
\caption{Human-label decoding accuracy. Model results are mean $\pm$
standard deviation over ten seeds.}
\label{tab:app-continuity-decoding}
\begin{tabular}{@{}lc@{}}
\toprule
Representation & Decoding accuracy \\
\midrule
Continuous MCD & $0.9615\pm0.0018$ \\
Hard $64$-state K-means & $0.5761\pm0.0099$ \\
Smoothed $64$-state K-means & $0.6216\pm0.0056$ \\
Majority-class baseline & $0.5566$ \\
\bottomrule
\end{tabular}
\end{table}

Smoothing improves the discrete baseline from $0.5761$ to $0.6216$, but
both remain far below continuous MCD ($0.9615$). Smooth transitions therefore
help, but do not recover most of the decoding performance lost through
this discretization.

\paragraph{Interpretation.}
The two analyses provide complementary evidence. Mixing persists within
stable annotated behaviors, and the continuous representation retains
behavioral information that the tested hard and smoothed discrete codes
lose. These results support graded, behaviorally informative variation
throughout behavior bouts, rather than mixing explained primarily by
smoothed transitions between discrete states.

\clearpage
\section{Hyperparameter sensitivity}

\subsection{Motif dimension}\label{app:ablation_motif_dimension}
To evaluate the sensitivity of our model to the dimension of the hidden state, we change the dimensions of motifs $d$ and verify them in different datasets.

\begin{figure*}[ht]
    \centering
    \includegraphics[width=1\textwidth]{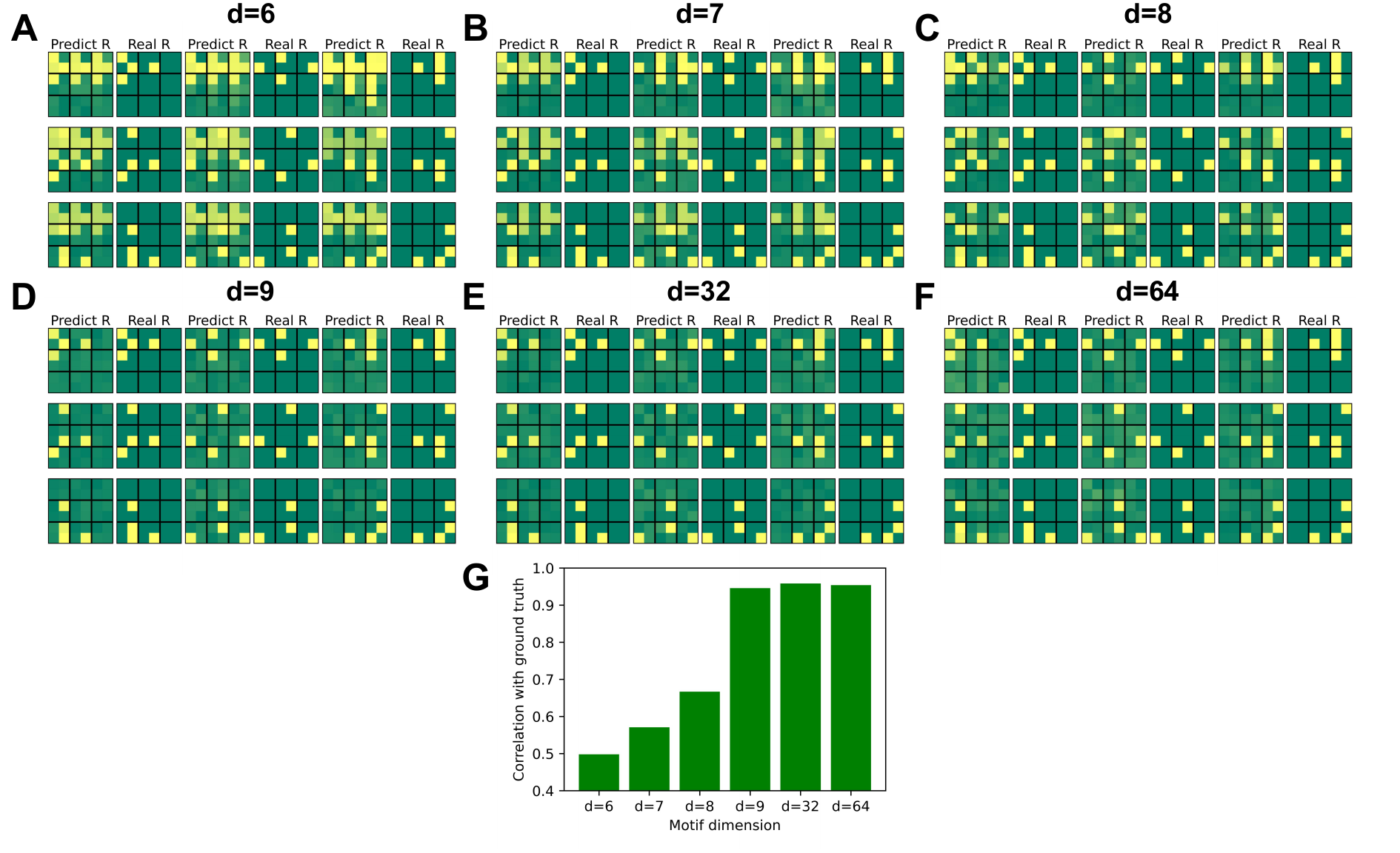}
  \caption{Reward maps generated by models of different motif dimension $d$ in the gridworld dataset.}
  \label{fig:app_gridworld_dimension}
\end{figure*}

\begin{figure*}[ht]
    \centering
    \includegraphics[width=1\textwidth]{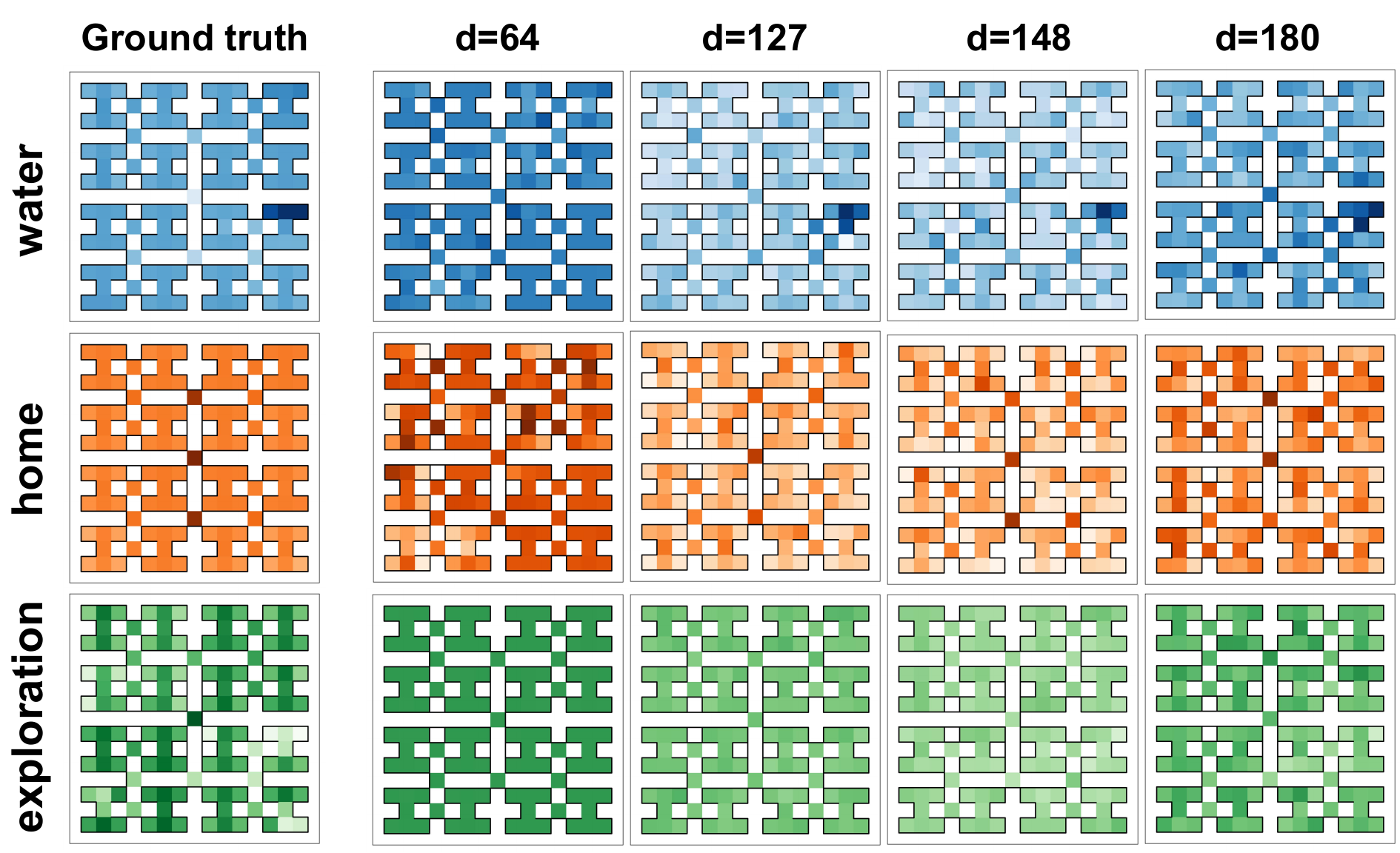}
  \caption{Reward maps generated by models of different motif dimension $d$ in the labyrinth navigation dataset.}
  \label{fig:app_labyrinth_dimension}
\end{figure*}

In the discrete case, there exists a minimum dimension that could fully represent the motif space. In fact, the motif discovery (Eq.~\ref{eq:p_linear}) is equivalent to SVD of the transition matrix $P(s'|s,a)$\citep{ren2022spectral}. Since the rank of the transition matrix is $|\mathcal{S}|$, the ranks of $\phi$ and $\mu$ matrices are both $|\mathcal{S}|$, which is exactly the minimum dimension of $\phi(s,a)$ that could fully represent the motif space. This could be clearly seen in the gridworld dataset (Fig.~\ref{fig:app_gridworld_dimension}) and labyrinth navigation dataset (Fig.~\ref{fig:app_labyrinth_dimension}). 

For the gridworld dataset, $d_{min}=3*3=9$. If $d<9$, the SVD could only provide a low-rank approximation. As a result, the recovered reward map is blurred and incorrect (Fig.~\ref{fig:app_gridworld_dimension}A-C) and the correlation remains low (Fig.~\ref{fig:app_gridworld_dimension}G). For $d\geq9$, the motif matrix is full-rank. So the reward maps are nearly perfect (Fig.~\ref{fig:app_gridworld_dimension}D-F), and the correlations are near 100\% (Fig.~\ref{fig:app_gridworld_dimension}G). Similar analysis also applies to labyrinth navigation dataset where $d_{min}=127$(Fig.~\ref{fig:app_labyrinth_dimension}). In this case, variant with fewer motif dimensions ($d=64$) cannot generate realistic reward maps while variants with more motif dimensions ($d\geq127$) can faithfully recover the reward peaks in each task.

\begin{figure*}[ht]
    \centering
    \includegraphics[width=0.8\textwidth]{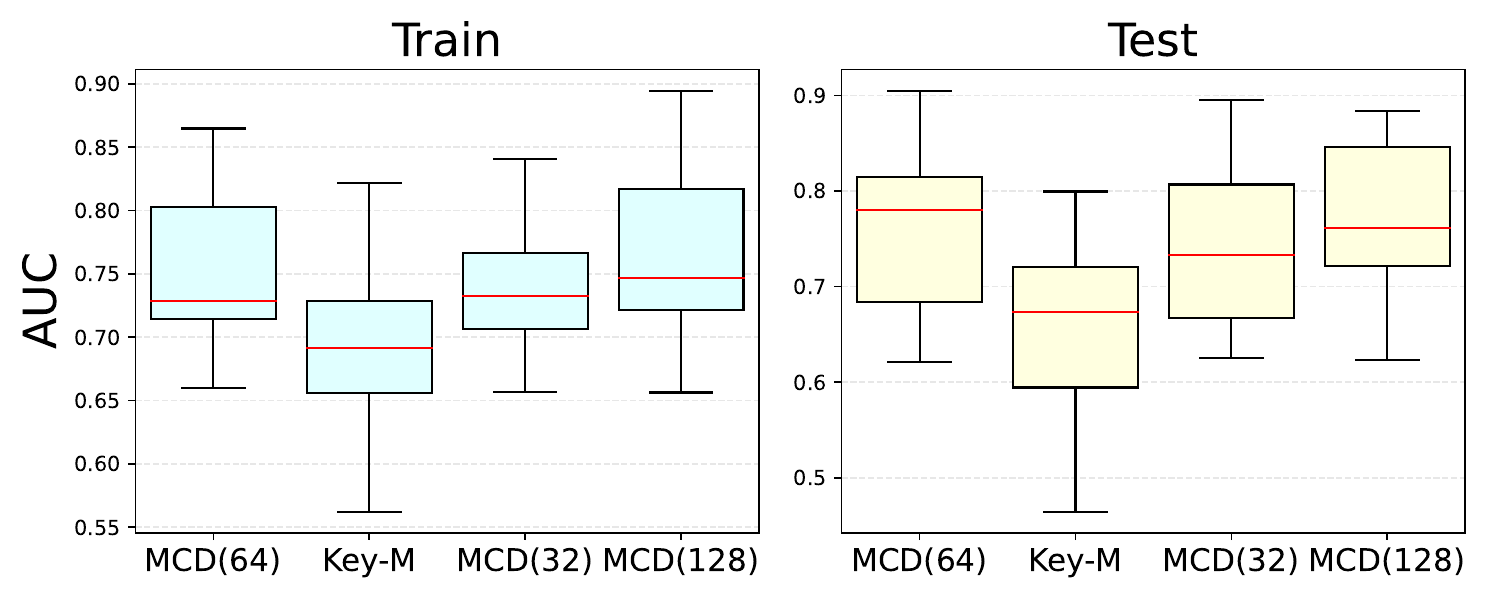}
  \caption{AUC generated by models of different motif dimension $d$ in the animal free-moving dataset.}
  \label{fig:app_keymoseq_dimension}
\end{figure*}

In the continuous case, since the transition kernel is infinite-dimension, any finite $d$ would only provide a low-rank approximation to the transition kernel. As $d$ increases, the fitting performance would be better. However, since the importance of different motifs is different, by observing the performance under different $d$, we could select one to be as small as possible while still capturing the entire motif space as comprehensively as possible, and maintain the performance. In this experiment (Fig.~\ref{fig:app_keymoseq_dimension}), we select $d=64$. 

\subsection{Noise distribution}\label{app:ablation_noise}
The continuous version of MCD is based on contrastive learning. Therefore, a high-quality negative sample distribution is crucial for the motif and policy learning. Here we evaluate different selections of noise distributions in the continuous free-moving animal behavior dataset. 

For the motif learning (Eq.~\ref{eq:NCEP}), in the main text, we directly sample states from the dataset. In other words, the noise distribution is $\rho(s)=\int\tau^e(s,a)da$. 

\begin{figure*}[ht]
    \centering
    \includegraphics[width=0.8\textwidth]{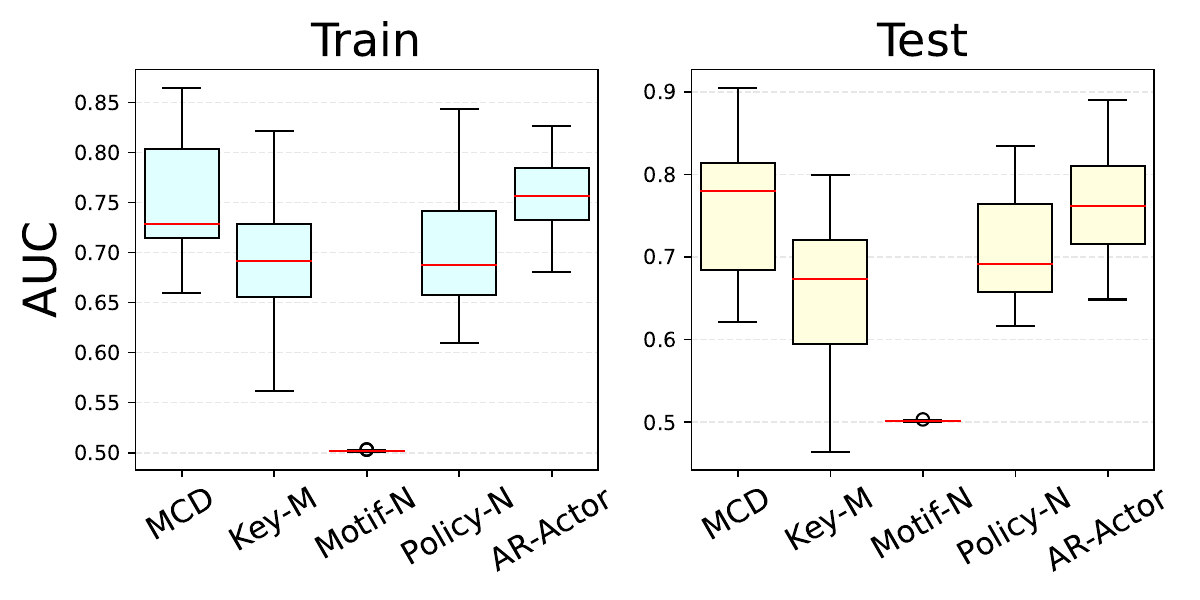}
  \caption{AUC generated by models trained under different noise distribution in the animal free-moving dataset.}
  \label{fig:app_keymoseq_noise}
\end{figure*}

In "Motif-N" variant, we replace it by a uniform distribution, whose dimension-wise bounds are determined as the maximum and minimum values of the states in the dataset, i.e. $s'_i\sim U(\min(s_i), \max(s_i))$. Because the negative samples are not good enough, the resulting energy-based model can neither capture high-quality motifs nor provide good basis vectors for later policy learning. Therefore the AUC score remains low (Fig~\ref{fig:app_keymoseq_noise}).

For the policy learning (Eq.~\ref{eq:NCEu}), in the main text, we also sample negative samples directly from the dataset $(s',a')\sim\tau^e$ and use $a'$ as a negative sample of actions. In other words, the noise distribution is $\zeta(a)=\int\tau^e(s,a)ds$.

In the "Policy-N" variant, we sample negative pairs from uniform distribution $a'_i\sim U(\min(\zeta(a_i)), \max(\zeta(a_i)))$, similar to "Motif-N".
In the "AR-Actor" variant, we additionally  train an auxilliary autoregressive actor $\pi(a|s)$ to fit $\pi^e(a|s)$ using MLE. And the negative samples are from this actor $\pi(a|s)$. For each dimension, the actor has a network of two hidden layers, with hidden dimension=256 each, to produce the mean and variance of this action dimension, i.e. $a_i\sim\mathcal{N}(\mu(a_i|s,a_{<i}),\sigma^2(a_i|s,a_{<i}))$. As expected, uniform noise distribution is not good enough, and "Policy-N" achieves lower AUC than the original version. And the "AR-Actor" achieves similar performance on the test dataset, proving the robustness of MCD to noise distribution if it is good enough. 

\begin{figure*}[ht]
    \centering
    \includegraphics[width=\textwidth]{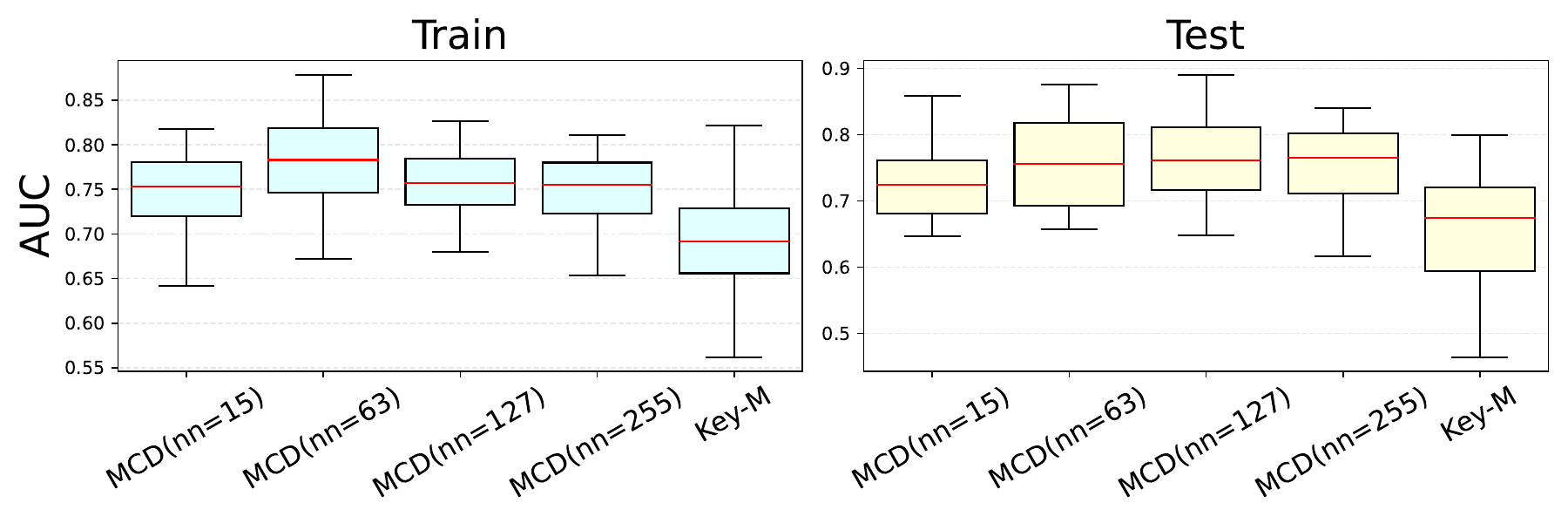}
  \caption{AUC generated by models trained using different numbers of negative samples for motif contrastive learning in the animal free-moving dataset. $nn=k$ means the \textbf{n}umber of \textbf{n}egative samples per positive sample is $k$.}
  \label{fig:app_keymoseq_n_neg}
\end{figure*}

We also tested the influences of the number of negative samples (Fig.~\ref{fig:app_keymoseq_n_neg}). MCD is consistently better than Key-Moseq and maintains a stable performance. It turns out that the model is robust against the choice of number of negative samples. 

\subsection{Dataset partition}\label{appendix:heldout}

\begin{figure*}[ht]
    \centering
    \includegraphics[width=0.5\textwidth]{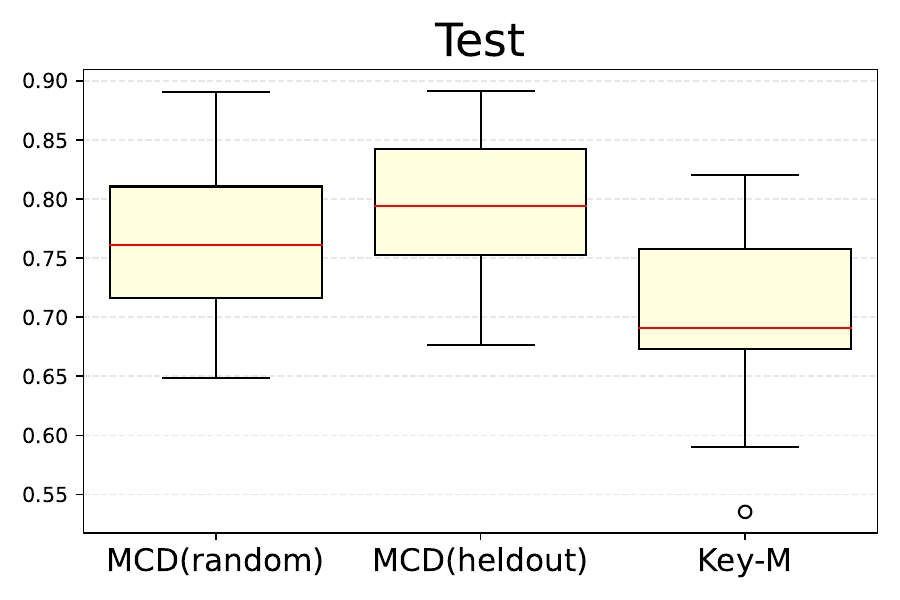}
  \caption{AUC generated by models trained using different train-test split methods. For MCD(random), the dataset is randomly split. For MCD(heldout), the behavior of one mouse is specially held out as the test set while the rest is taken as the training set.}
  \label{fig:app_keymoseq_heldout}
\end{figure*}

In the third experiment, we randomly split the whole dataset into training set and test set. Here, we train the model on some mice and then test it on other heldout mice, i.e. split the dataset by mouse identities. This experiment shows whether the motifs learned by the model are general and transferrable across different subjects. The result is shown in Fig.~\ref{fig:app_keymoseq_heldout} where the complete exclusion of one animal from the training set does not impair the performance. This experiment proves that the MCD can learn a general set of motifs that could be transferred across animals.

\subsection{Gaussian random walk prior}\label{appendix:grw_prior}

\begin{figure*}[ht]
    \centering
    \includegraphics[width=0.8\textwidth]{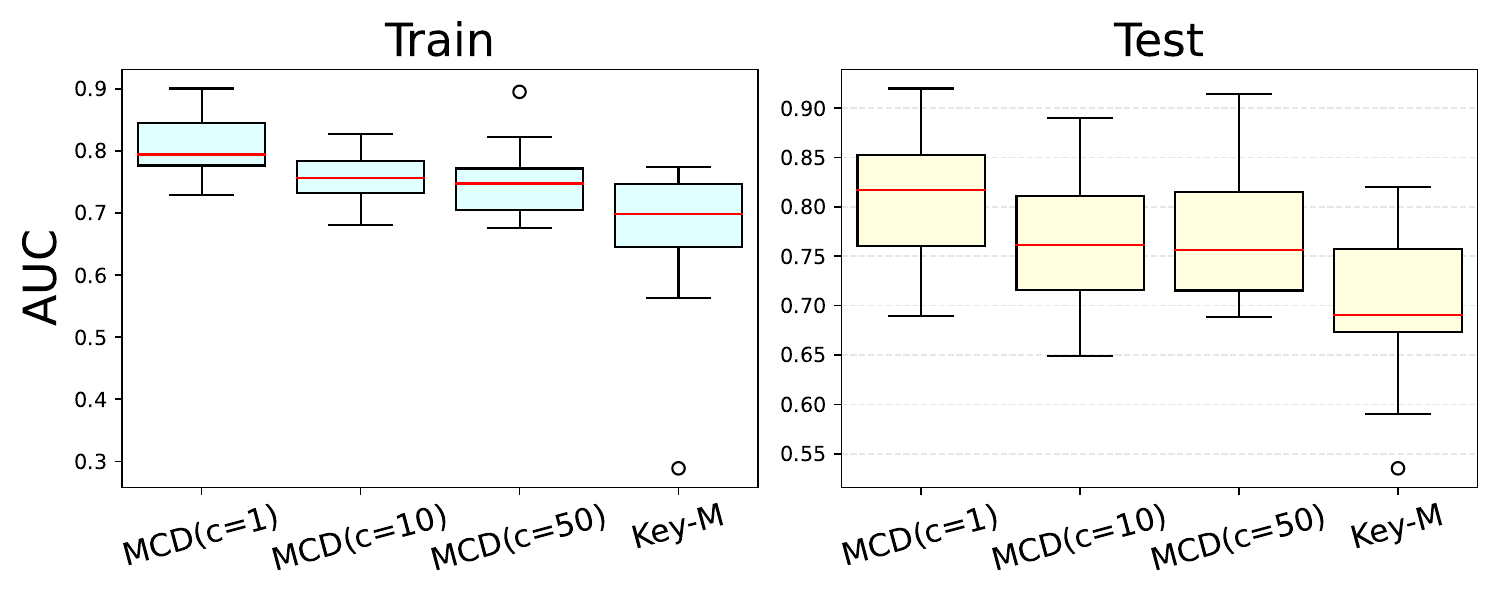}
  \caption{AUC generated by models tested using different Gaussian random walk prior coefficients in the animal free-moving dataset. }
  \label{fig:app_keymoseq_grw}
\end{figure*}

In the free-moving animal behavior dataset (Sec.~\ref{sec:exp3}), we impose a Gaussian random-walk prior on the time-varying motif weights $u(t)$ to encourage temporal smoothness. To assess the sensitivity of our method to this regularization, we varied the strength of the Gaussian prior and re-evaluated model performance (Fig.~\ref{fig:app_keymoseq_grw}). As expected, very large coefficients introduce slight degradation due to oversmoothing. However, across all tested values, performance remains substantially higher than the baseline model. This demonstrates that MCD is robust to the choice of Gaussian random-walk regularization strength and does not rely on fine-tuning this hyperparameter to achieve strong results.

\subsection{Temporal resolution}\label{appendix:temporal_resolution}

\begin{figure*}[ht]
    \centering
    \includegraphics[width=\textwidth]{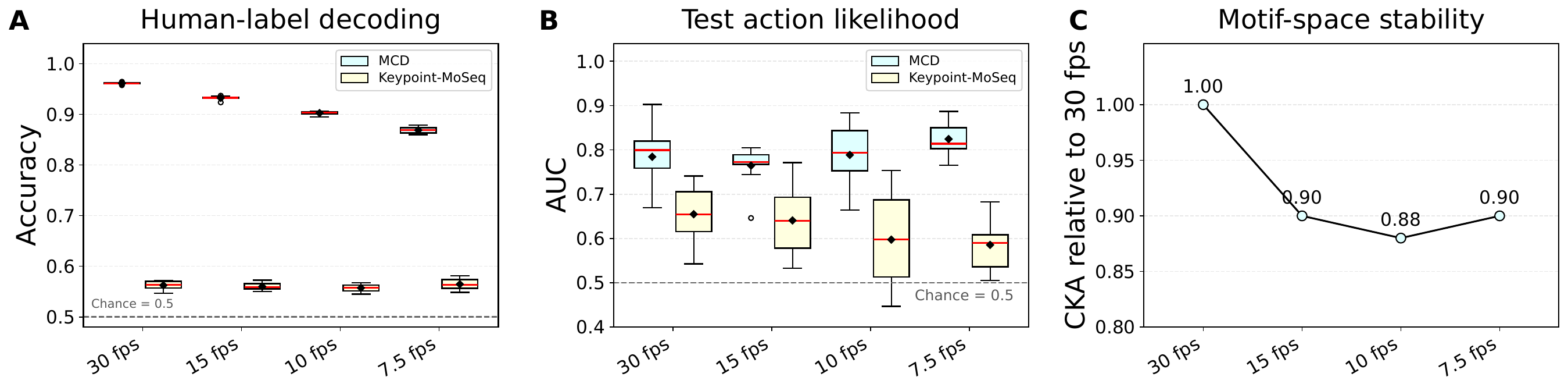}
  \caption{Decoding accuracy (\textbf{A}) and AUC (\textbf{B}) generated by models tested using different downsampling rates in the animal free-moving dataset. \textbf{C.} CKA under different downsampling rates for MCD in the same dataset.}
  \label{fig:app_temporal_resolution}
\end{figure*}

We evaluate the sensitivity of MCD to temporal resolution by downsampling the original 30 fps videos to 15, 10, and 7.5 fps, with actions recomputed between new frames. We then use the same MCD checkpoint trained at 30 fps to test their performance (Fig.~\ref{fig:app_temporal_resolution}). As in the main text, we report two metrics. \textbf{Human-label decoding accuracy} tests whether the fitted MCD representation preserves behaviorally meaningful information. \textbf{Test action-likelihood AUC} measures whether MCD assigns higher likelihood to the observed action than to mismatched actions. Chance AUC is 0.5. We also use \textbf{Motif space CKA} to measure how different the motif spaces are across downsampling rates.

Human-label decoding accuracy decreases gradually as temporal resolution is reduced, while remaining competitive with other baselines. In contrast, test action-likelihood AUC remains well above chance and shows no degradation. The CKA shows only minor decrease, suggesting that the motif space is stable across sampling rates and captures action-relevant information. Therefore, MCD is robust to temporal resolution.\label{appendix:sensitivity}

\clearpage
\section{Generated rollout trajectories}

\begin{figure*}[ht]
    \centering
    \includegraphics[width=\textwidth]{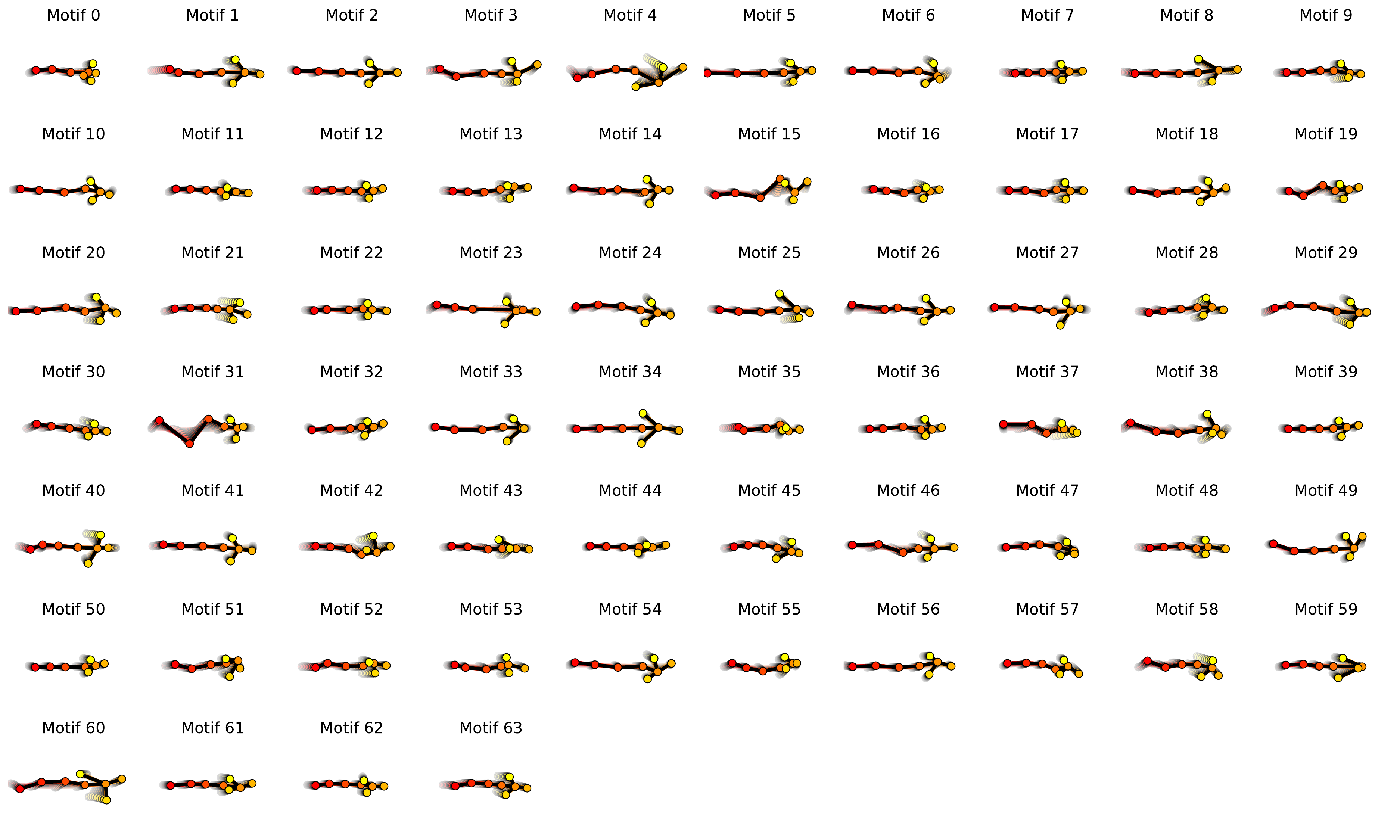}
  \caption{Motif-specific rollout trajectories generated by HMC sampling from the trained energy model. Indexes are consistent with Fig.~\ref{fig:app_keymoseq}.}
  \label{fig:app_keymoseq_rollout}
\end{figure*}

In the animal free-moving dataset, we train a generative energy model to estimate the state--action value function. Here, we assess its reliability using 
Hamiltonian Monte Carlo (HMC) sampling. For each motif $i\in{1,...,d}$, we construct a one-hot vector $u$ with $u_{i}=1$ and $u_{j}=0$ for all $j\neq i$, 
and then sample an action $a$ from the learned energy model. For each motif, we sample $10$ actions in total. The resulting motif-specific rollout trajectories (Fig~\ref{fig:app_keymoseq_rollout}) largely align with the empirical averages (Fig.~\ref{fig:app_keymoseq}B), demonstrating that the model captures the dominant behavioral tendencies. A small number of trajectories deviate from the empirical trends, probably because $u$ is never used as a strict one-hot vector during training but instead appears as a mixture over motifs. We attribute these deviations to unavoidable generative noise and the distributional mismatch between test-time one-hot inputs and the mixed representations observed in real data. 

\begin{figure*}[ht]
    \centering
    \includegraphics[width=\textwidth]{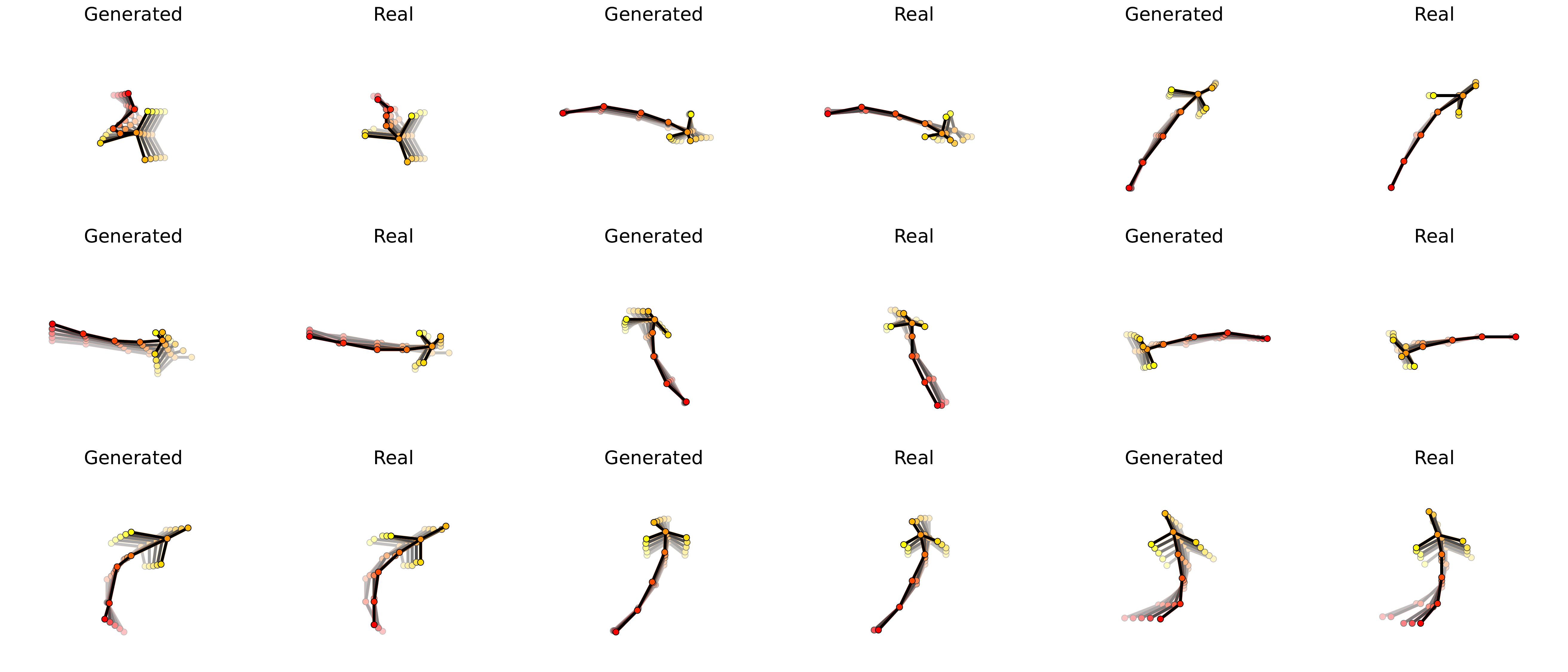}
  \caption{Rollout trajectories based on inferred $u(t)$ generated by HMC sampling from the trained energy model. }
  \label{fig:app_keymoseq_rollout_from_idx}
\end{figure*}

For that, we choose several clips and generate the rollouts (number of rollout steps=5) based on the inferred $u(t)$. The alignment between generated trajectories and real trajectories (Fig.~\ref{fig:app_keymoseq_rollout_from_idx}) show that MCD could capture the real behavior dynamics. 

\paragraph{HMC parameters.} We use a step size of $1\times10^{-4}$, 100 leapfrog steps per action, temperature $=1$, mass matrix $M=I$, and initial momentum $r\sim\mathcal N(0, 1/200)$. The leapfrog integrator is employed to improve sampling accuracy.

\label{appendix:rollout}
\label{sec:appendix}

\clearpage
\section*{NeurIPS Paper Checklist}

\begin{enumerate}

\item {\bf Claims}
    \item[] Question: Do the main claims made in the abstract and introduction accurately reflect the paper's contributions and scope?
    \item[] Answer: \answerYes{} 
    \item[] Justification: In Sec.~\ref{sec:skil}, building on results in reinforcement learning and imitation learning, we propose the theoretical framework of our model MCD for behavior understanding. In Sec.~\ref{sec:exp}, we apply our methods to three different datasets: simulated grid world trajectories, trajectories of a real mouse navigating in a labyrinth, the poses of a freely moving mouse, and macaque drawing behavior. We present the interpretation of our results and show that the performance of our results is better than similar models.
    

\item {\bf Limitations}
    \item[] Question: Does the paper discuss the limitations of the work performed by the authors?
    \item[] Answer: \answerYes{} 
    \item[] Justification:  The paper explicitly discusses limitations in Sec.~\ref{sec:discuss}, including sensitivity to input data quality (e.g., tracking errors) and challenges in validating correspondence between learned motifs and neural mechanisms. 

\item {\bf Theory assumptions and proofs}
    \item[] Question: For each theoretical result, does the paper provide the full set of assumptions and a complete (and correct) proof?
    \item[] Answer: \answerYes{} 
    \item[] Justification: We define motif sets in Definition~\ref{def:motif_set}, and combine it with the reinforcement learning framework to describe the learning procedure of skill sets $\phi(s,a)$ and motif weights $u(t)$. In Sec.~\ref{sec:continuous_version}, we solve the difficulty in the continuous dataset using energy-based model (EBM) and prove the connection between EBM and our definition of motif sets in Appendix~\ref{sec:equivalence}. We also derive how to infer reward functions under our framework in Sec.~\ref{sec:rewardrecovery}. 

    \item {\bf Experimental result reproducibility}
    \item[] Question: Does the paper fully disclose all the information needed to reproduce the main experimental results of the paper to the extent that it affects the main claims and/or conclusions of the paper (regardless of whether the code and data are provided or not)?
    \item[] Answer: \answerYes{} 
    \item[] Justification: The paper provides sufficient details to reproduce the main experimental results, including theoretical formulations, model architectures, training objectives, dataset construction, evaluation protocols, hyperparameter settings, and compute resources in the main text and appendix. In addition, the supplementary material includes the implementation code used for training and evaluation. 

\item {\bf Open access to data and code}
    \item[] Question: Does the paper provide open access to the data and code, with sufficient instructions to faithfully reproduce the main experimental results, as described in supplemental material?
    \item[] Answer: \answerYes{} 
    \item[] Justification: We provide our code in the supplemental material as well as the code link. 

\item {\bf Experimental setting/details}
    \item[] Question: Does the paper specify all the training and test details (e.g., data splits, hyperparameters, how they were chosen, type of optimizer) necessary to understand the results?
    \item[] Answer: \answerYes{} 
    \item[] Justification: The paper provides sufficient experimental details to understand and reproduce the results, including dataset construction and evaluation protocols in Sec.~\ref{sec:exp}, as well as comprehensive hyperparameter settings, model architectures, and training procedures in Appendix~\ref{appendix:hyperparameter}. Additionally, the code and preprocessing details are included in the supplementary material.
    

\item {\bf Experiment statistical significance}
    \item[] Question: Does the paper report error bars suitably and correctly defined or other appropriate information about the statistical significance of the experiments?
    \item[] Answer: \answerYes{} 
    \item[] Justification: In Fig.~\ref{fig:exp3}, we use boxplots to summarize variability in human-annotated label prediction accuracy and AUC across test evaluations. To support the main performance comparison, we additionally perform paired t-tests against each baseline and report the resulting $p$-values in Sec.~\ref{sec:exp3}.
    

\item {\bf Experiments compute resources}
    \item[] Question: For each experiment, does the paper provide sufficient information on the computer resources (type of compute workers, memory, time of execution) needed to reproduce the experiments?
    \item[] Answer: \answerYes{} 
    \item[] Justification: The paper specifies compute resources in Appendix~\ref{sec:appendix_compute_resource}, including hardware (CPU, GPU, memory), software framework, and per-experiment training time across all datasets. 
    
\item {\bf Code of ethics}
    \item[] Question: Does the research conducted in the paper conform, in every respect, with the NeurIPS Code of Ethics \url{https://neurips.cc/public/EthicsGuidelines}?
    \item[] Answer: \answerYes{} 
    \item[] Justification: We have reviewed the NeurIPS Code of Ethics and confirm that our work complies with its requirements. Our research uses publicly available or standard behavioral datasets without personal identifiable information, does not involve human subjects or sensitive data collection, and poses minimal risk of harmful application. Potential societal impacts, including both benefits and risks, are discussed in the Impact statement (Appendix~\ref{sec:impact_statement}), and no aspect of the work relates to disallowed areas such as surveillance, discrimination, or weaponization. 

\item {\bf Broader impacts}
    \item[] Question: Does the paper discuss both potential positive societal impacts and negative societal impacts of the work performed?
    \item[] Answer: \answerYes{} 
    \item[] Justification: The paper discusses broader impacts in the Impact statement (Appendix~\ref{sec:impact_statement}), including positive contributions such as advancing understanding of animal behavior and potential applications in neuroscience, as well as negative risks such as possible misuse when extending the framework to model human behavior (e.g., bias amplification). These considerations provide a balanced discussion of both beneficial and harmful societal implications.
    
\item {\bf Safeguards}
    \item[] Question: Does the paper describe safeguards that have been put in place for responsible release of data or models that have a high risk for misuse (e.g., pre-trained language models, image generators, or scraped datasets)?
    \item[] Answer: \answerNA{} 
    \item[] Justification: The paper does not release models or datasets that pose a high risk of misuse (e.g., large pre-trained models, generative systems, or sensitive datasets). Our work focuses on methodological contributions evaluated on controlled and standard behavioral datasets, and therefore does not require specific safeguards for responsible release.

\item {\bf Licenses for existing assets}
    \item[] Question: Are the creators or original owners of assets (e.g., code, data, models), used in the paper, properly credited and are the license and terms of use explicitly mentioned and properly respected?
    \item[] Answer: \answerYes{} 
    \item[] Justification: All datasets, environments, and baseline methods used in this work are properly credited through citations in the main text and references. Their licenses and terms of use, including the Harvard non-commercial academic license for Keypoint-MoSeq and the MIT licenses for the labyrinth dataset and Gymnasium environments, are explicitly documented in Appendix~\ref{sec:appendix_license}.

\item {\bf New assets}
    \item[] Question: Are new assets introduced in the paper well documented and is the documentation provided alongside the assets?
    \item[] Answer: \answerYes{} 
    \item[] Justification: The paper introduces new code implementing the proposed MCD framework, which is provided in the supplementary material and the link. We document training procedures, model architectures, and hyperparameters in the appendix, enabling reproducibility and proper use of the released assets.

\item {\bf Crowdsourcing and research with human subjects}
    \item[] Question: For crowdsourcing experiments and research with human subjects, does the paper include the full text of instructions given to participants and screenshots, if applicable, as well as details about compensation (if any)? 
    \item[] Answer: \answerNA{} 
    \item[] Justification: The paper does not involve crowdsourcing or research with human subjects. All experiments are conducted on simulated environments and existing animal behavior datasets.

\item {\bf Institutional review board (IRB) approvals or equivalent for research with human subjects}
    \item[] Question: Does the paper describe potential risks incurred by study participants, whether such risks were disclosed to the subjects, and whether Institutional Review Board (IRB) approvals (or an equivalent approval/review based on the requirements of your country or institution) were obtained?
    \item[] Answer: \answerNA{} 
    \item[] Justification:  The paper does not involve research with human subjects or crowdsourcing. All experiments are conducted on simulated environments and existing animal behavior datasets, and therefore IRB approval is not applicable.

\item {\bf Declaration of LLM usage}
    \item[] Question: Does the paper describe the usage of LLMs if it is an important, original, or non-standard component of the core methods in this research? Note that if the LLM is used only for writing, editing, or formatting purposes and does \emph{not} impact the core methodology, scientific rigor, or originality of the research, declaration is not required.
    \item[] Answer: \answerNA{} 
    \item[] Justification:  The core methodology of this work does not involve LLMs. While a language model was used only for writing assistance (e.g., grammar and clarity), it did not contribute to the scientific content, model design, or experimental results.

\end{enumerate}

\end{document}